# Benchmarking Denoising Algorithms with Real Photographs

Tobias Plötz Stefan Roth
Department of Computer Science, TU Darmstadt

## Abstract

*Lacking realistic ground truth data, image denoising techniques are traditionally evaluated on images corrupted by synthesized* i. i. d. *Gaussian noise. We aim to obviate this unrealistic setting by developing a methodology for benchmarking denoising techniques on real photographs. We capture pairs of images with different ISO values and appropriately adjusted exposure times, where the nearly noise-free low-ISO image serves as reference. To derive the ground truth, careful post-processing is needed. We correct spatial misalignment, cope with inaccuracies in the exposure parameters through a linear intensity transform based on a novel heteroscedastic Tobit regression model, and remove residual low-frequency bias that stems,* e.g.*, from minor illumination changes. We then capture a novel benchmark dataset, the* Darmstadt Noise Dataset (DND)*, with consumer cameras of differing sensor sizes. One interesting finding is that various recent techniques that perform well on synthetic noise are clearly outperformed by BM3D on photographs with real noise. Our benchmark delineates realistic evaluation scenarios that deviate strongly from those commonly used in the scientific literature.*

## 1. Introduction

Noise is inherent to every imaging system. Especially in low-light scenarios, it often severely degrades the image. Therefore, a large variety of denoising algorithms have been developed to deal with image noise, *e.g.* [4, 5, 6, 7, 8, 25, 31, 32, 39]. Even though images with real sensor noise can be captured easily, it is much less straightforward to know what the true noise-free image should be. Thus, the quantitative evaluation of denoising methods by and large relies on adding synthetic *i. i. d.* Gaussian noise to mostly clean images, *e.g.* [19, 31, 32]. Photographs with real noise are at best used for a qualitative analysis [25, 29], but often not at all. This is quite problematic, since noise in real photographs is not *i. i. d.* Gaussian [13, 21], yet even seemingly minor details of the synthetic noise process, such as whether the noisy values are rounded to integers, can have a significant effect on the relative performance of methods [6, 33].

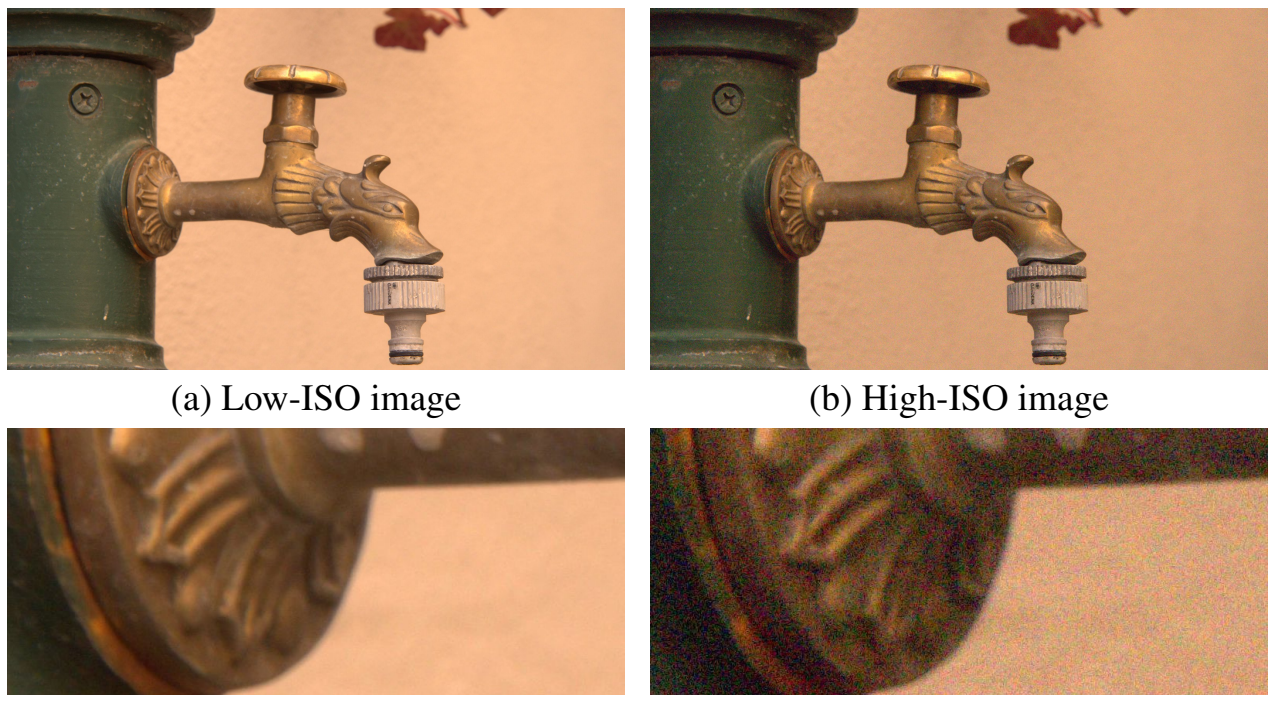

(a) Low-ISO image (b) High-ISO image
(c) Zoom-in of the low-ISO image (left) and the high-ISO image (right)

Figure 1. An image pair of a nearly noise-free low-ISO and a noisy high-ISO image from our dataset. Note, that we work with RAW images and show JPEGs for better display.

The goal of this paper is to address these challenges by developing a methodology for benchmarking denoising algorithms by means of real photographs. At its core is the simple idea of capturing pairs of noisy and almost noise-free images by imaging the same scene from the same viewpoint with different analog gains (ISO values), see Fig. 1. By inversely adjusting the exposure time, the underlying noise-free image intensities should theoretically stay constant. In practice, we observe various causes for changing image intensities, prohibiting the direct use of the low-ISO image as ground truth. Since these scene variations are nontrivial, we contribute a careful *post-processing procedure* that takes into account the statistical properties of the image formation process. As part of this pipeline we propose a novel *heteroscedastic Tobit regression model* generalizing [35], which allows to remove linear dependencies between the intensities of both images that arise as neither the analog gain of the sensor nor the exposure time can be controlled completely accurately in practice. Our model faithfully accounts for clipping as well as signal-dependent noise, which is crucial as shown experimentally. Furthermore, minimal changes in the illumination can lead to a low-frequency bias, which we remove by high-pass filtering the residual between noisy and reference image in a transformed domain in which the noise process is zero-mean. Lastly, moving objects and minuscule camera shake between exposures

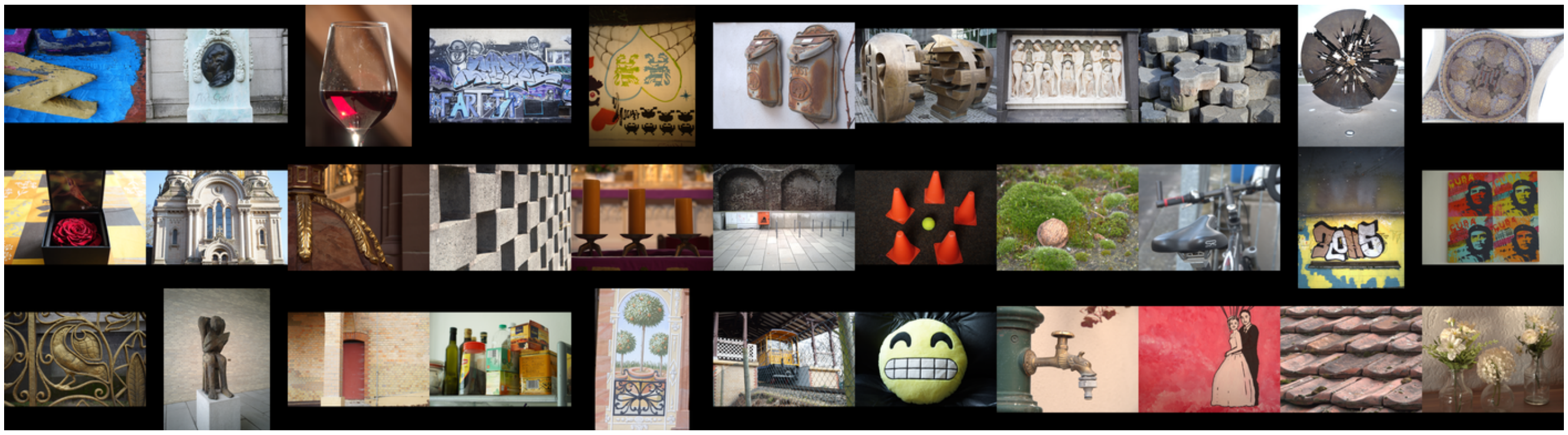

Figure 2. An overview of the scenes used in our benchmark dataset (subset shown).

are treated by manual annotation and simple Lucas-Kanade subpixel alignment [23], respectively.

Based on this acquisition pipeline, we capture a *real-world dataset* of image noise, called Darmstadt Noise Dataset (DND). We use 4 consumer cameras, ranging from a smartphone with a $^1/_{2.3}$ inch sensor to a full-frame interchangeable lens camera. Images are taken across a wide range of different ISO values in realistic photographic situations, providing a novel reference dataset for benchmarking denoising algorithms in realistic conditions. Our dataset consists of 50 scenes and is publicly available.[1] Figure 2 shows a subset of the scenes.

Our realistic dataset enables *interesting insights* into the performance of recent denoising algorithms. We find that a number of current techniques (*e.g.*, NCSR [8], WNMM [15], TNRD [6]) that – based on previous analyses with synthetic *i. i. d.* Gaussian noise – were presumed to outperform the by now classic BM3D [7], do in fact perform worse than BM3D on photographs with real noise. Moreover, our analysis reveals that noise strengths for consumer cameras are significantly lower than what is usually assumed in the scientific literature when evaluating denoising algorithms. We further highlight the importance of applying denoising before the non-linear camera processing pipeline [30]. Our findings strongly question the *practical relevance* of previous synthetic evaluation methodologies.

## 2. Related Work

Since noise is abundant in any imaging system, its statistical properties have been well studied. Thorough analyses have been provided for CCD [18] and CMOS image sensors [9]. One inevitable source of noise is induced by the stochastic arrival process of photons hitting the sensor – so-called shot noise. Since it follows a Poisson distribution, its variance is proportional to the mean intensity at a specific pixel and is hence not stationary across the whole image. Other noise sources originate from the electronics within the sensor chip and from discretization [9, 13, 18].

[1] https://noise.visinf.tu-darmstadt.de

Although the image noise variance depends on the underlying intensity, the majority of denoising algorithms ignore this and evaluates against artificial, stationary noise, usually assumed *i. i. d.* Gaussian, *e.g.* [31, 32, 39]. Other works specifically aim to *model intensity-dependent noise* [21, 24]. The main idea there is to model the noise distribution as a heteroscedastic Gaussian, whose variance is intensity-dependent. This is valid since the Poissonian components of the total noise can be approximated well with a Gaussian. Other approaches first apply a variance stabilizing transform [11, 26] and subsequently employ a denoising method for stationary Gaussian noise. However, the transform may make the noise distribution non-Gaussian [37].

There have been attempts to validate denoising algorithms on real data at a small scale [22, 38]. They rely on recovering a noise-free image by temporal averaging several noisy observations. However, they ignore the fact that the noise process is not zero-mean due to clipping effects [11], whereas we show that it is important to consider this bias when creating a denoising ground truth. They also do not take potentially further non-linear processing of raw intensities [20] into account.

To the best of our knowledge, the only other effort on benchmarking denoising with real images is the RENOIR dataset [2]. It also relies on taking sets of images of a static scene with different ISO values, but the post-processing is less refined. Image pairs appear to exhibit spatial misalignment, the intensity transform does not model heteroscedastic noise, and low-frequency bias is not removed. Our experiments indicate that ignoring these sources of error significantly affects the realism of the dataset. Moreover, [2] is based on 8 bit demosaiced images while we work with untainted linear raw intensities.

It is often useful to measure the noise characteristics of a sensor at a certain ISO level. [10] proposes to illuminate the sensor with approximately constant irradiation and subsequently aggregates intensity measurements *spatially*. This is repeated for different irradiation levels to capture the intensity dependence of the noise. [12, 28] propose a less tedious capture protocol similar to ours, where multiple exposures of a static scene are used to aggregate the measurements at every pixel site *temporally*. In contrast, our Tobit regression allows to estimate the parameters of the noise process by having access to just two images.

## 3. Image Model and Data Acquisition

We begin by motivating the capture protocol and post-processing for our dataset; we also detail the image formation process underlying the data acquisition, see Fig. 3.

**Image formation.** Capturing a noisy image $x_n$ can be described by adding noise to a latent noise-free image $y_n$ and afterwards clipping the intensities to account for the saturation of pixels on the sensor:

$$x_n = \text{clip}\big(y_n + \epsilon_n(y_n)\big), \tag{1}$$

where $\text{clip}(y) = \min(\max(y, 0), 1)$ and $\epsilon_n$ can be modeled as Poisson-Gaussian noise whose strength depends on the noise-free intensity. Following [3, 11], we approximate the noise distribution with a heteroscedastic Gaussian

$$\epsilon_n(y_n) \sim \mathcal{N}\big(0, \sigma_n(y_n)\big) \tag{2a}$$

$$\text{with} \quad \sigma_n^2(y_n) = \beta_1^n y_n + \beta_2^n, \tag{2b}$$

where $\sigma_n(y_n)$ is called the *noise level function* with parameters $\boldsymbol{\beta}^n$. Due to the clipping, naïve temporal or spatial averaging of the noisy observations will yield a bias, *i.e.* $\mathbb{E}\left[x_n \mid y_n\right] \neq y_n$. However, we can express $\mathbb{E}\left[x_n \mid y_n\right]$ analytically in terms of $y_n$ and $\sigma_n(y_n)$, see [11] for details, and denote this relation as

$$\mathcal{A}(y_n) \doteq \mathbb{E}\left[x_n \mid y_n\right]. \tag{3}$$

Ideally, we would want to use $y_n$ as ground truth for denoising $x_n$. However, since $y_n$ is not available, we propose to take another picture $x_r$ that shows the same scene as $x_n$, but is affected only little by noise. Since the parameters $\boldsymbol{\beta}$ of the noise-level-function depend mainly on the camera sensor and on the ISO value [10], we achieve this by using a low ISO value to obtain the reference image $x_r$.

**Capture protocol and residual errors.** As this reference image $x_r$ is captured at a different time instant and with a different exposure time and ISO value than $x_n$, it is generated from a second latent image $y_r$ with noise parameters $\boldsymbol{\beta}^r$, analogously to Eq. (1). In practice, we take the reference at the base ISO level of the camera, while the ISO value for the noisy image is $n$ times larger. To compensate this, the reference image is taken with $n$ times the exposure time. All other camera parameters including aperture, white balance, and focus remain constant. Since the latent, noise-free image intensity is proportional to both the ISO value and the exposure time, *in theory* our capture protocol leaves the noise-free image intensities invariant, *i.e.* $y_n = y_r$. As $x_r$ exhibits only very little noise, *i.e.* $x_r \approx y_r$, we could use $x_r$ instead of $y_n$ as denoising ground truth.

Table 1. Cameras used for capturing the dataset.

| Camera | # img. | Sensor size [mm] | Res. [Mpix] | ISO |
|---|---|---|---|---|
| Sony A7R | 13 | $36 \times 24$ | 36.3 | 100 – 25.6k |
| Olympus E-M10 | 13 | $17.3 \times 13$ | 16.1 | 200 – 25.6k |
| Sony RX100 IV | 12 | $13.2 \times 8.8$ | 20.1 | 125 – 8k |
| Huawei Nexus 6P | 12 | $6.17 \times 4.55$ | 12.3 | 100 – 6.4k |

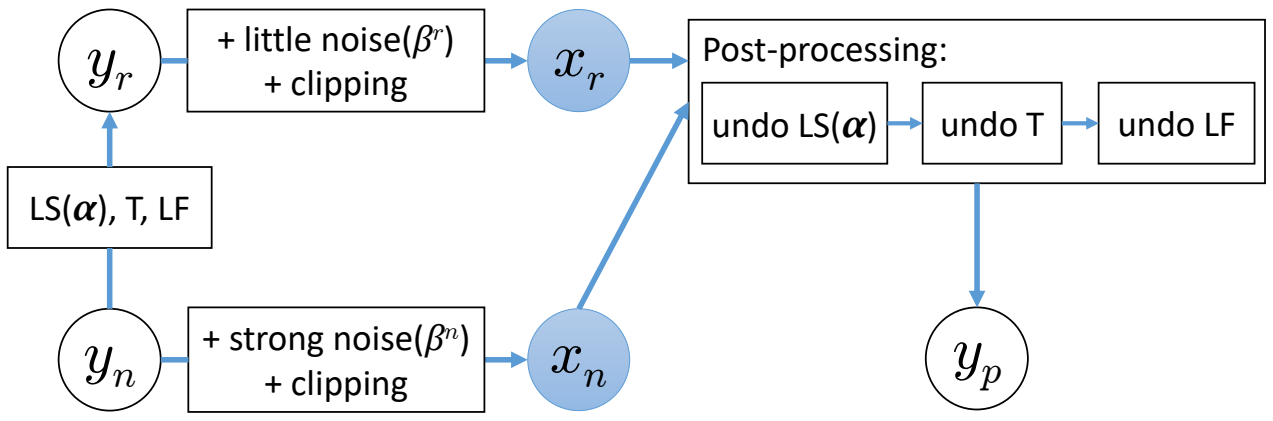

Figure 3. Image formation process underlying the observed low-ISO image $x_r$ and high-ISO image $x_n$. They are generated from latent noise-free images $y_r$ and $y_n$, respectively, which in turn are related by a linear scaling of image intensities (LS), a small camera translation (T), and a residual low-frequency pattern (LF). To obtain the denoising ground truth $y_p$, we apply post-processing to $x_r$ aiming at undoing these undesirable transformations.

For the noise-free intensity to truly stay the same, the captured scene and the camera have to be static and the illumination has to remain constant. Neither is generally the case. To minimize the effect of camera shake and scene variation during acquisition, we developed an Android app that quickly issues all necessary commands to the camera over WiFi. We mount the camera on a sturdy tripod with a stabilizing weight attached. Moreover, we use mirrorless cameras, which reduces vibrations due to mirror flapping compared to DSLRs. Despite this careful protocol, we still observe residual errors that we undo using the pipeline detailed in Sec. 4; post-processing $x_r$ results in a new image $y_p$. In Sec. 5 we show that $y_p$ is now sufficiently close to $y_n$ and hence use $y_p$ as ground truth for our benchmark.

**Further details.** For our image database described in Sec. 6 we use four different cameras, see Table 1. The cameras span a substantial range of sensor sizes from $1/2.3$ inch to a full-frame sensor. We extract linear raw intensities from the captured images using the free software *dcraw*. Afterwards we scale image intensities to fall inside the range $[0, 1]$ by normalizing with the black and white level.

## 4. Post-Processing

Our post-processing aims at undoing undesirable transformations between the latent images $y_n$ and $y_r$. These are revealed by looking at the difference images between the low-ISO image $x_r$ and high-ISO image $x_n$ (Fig. 4a).

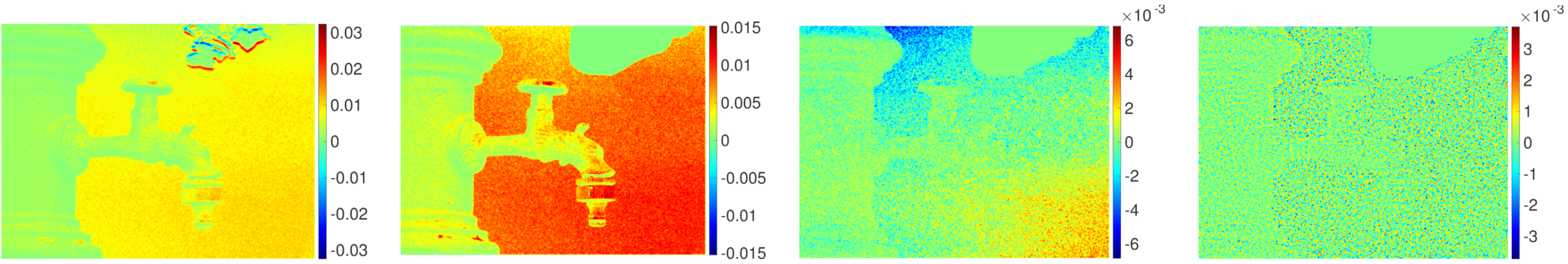

(a) Difference (residual) image $R(x_r)$ between Figs. 1a and 1b

(b) After manually masking moving objects

(c) Residual $R(y_p'')$ after linear intensity scaling and alignment

(d) Final residual $R(y_p)$ after subsequent removal of low-frequency residuals

Figure 4. Difference between blue channels of low- and high-ISO images from Fig. 1 after various post-processing stages. Images are smoothed *for display* to highlight structured residuals, attenuating the noise.

Specifically we consider the *debiased residual image* $R(x_r)$ with

$$R(\cdot) \doteq \mathcal{A}(\cdot) - x_n. \tag{4}$$

From Eq. (3) it immediately follows that the *ground truth debiased residual image* $R(y_n)$ is zero-mean. However, from Fig. 4a it is apparent that $R(x_r)$ is not zero-mean. We trace this to four sources of errors that need to be corrected for in order to relate the intensities of a certain pixel across the different exposures: *(i)* In general scenes individual objects may move during the capture procedure; *(ii)* spatial sub-pixel misalignments may be caused by small camera vibrations, *e.g.*, due to the mechanical shutter; *(iii)* the lighting of the scene may change slightly during capture, outdoors for example because of moving clouds, indoors for example due to light flicker; *(iv)* linear intensity changes arise from the fact that neither the analog gain nor the exposure time can be perfectly controlled. Note, that the severity of *(i)*–*(iii)* aggravates the more pictures are taken, thus complicating the use of temporal averaging methods for creating denoising ground truth in realistic scenes. Our capture protocol strikes a balance between *(i)*–*(iii)* and *(iv)* by requiring the minimum of only two exposures, while creating the need to account for linear intensity changes.

We need to cope with these four sources of errors to obtain an accurate ground truth. We address *(i)* by masking objects with a simple GUI tool. Our post-processing aims at undoing *(ii)*–*(iv)*, *c.f.* Fig. 3. We model *(ii)* as a global 2D translation and *(iv)* as a linear scaling of pixel intensities, both of which can be inverted given an estimate of their underlying parameters. Any remaining low-frequency bias *(iii)* is removed in a final filtering step, producing the post-processed image $y_p$. We now detail these steps.

**Linear intensity changes.** Changing the analog amplifier gain and the exposure time introduces a linear relationship between $y_n$ and $y_r$ (Fig. 4b), since neither of those parameters can be controlled with perfect accuracy:

$$y_n = \alpha(y_r) = \alpha_1 y_r + \alpha_2, \tag{5}$$

where the offset $\alpha_2$ accounts for inaccuracies of the recorded black level. As we do not have access to $y_n$ and $y_r$, we need to estimate $\alpha_1, \alpha_2$ from the observed images. From Eq. (1), we relate $x_r$ and $x_n$ as

$$\begin{aligned} x_n &= \operatorname{clip}\big(y_n + \epsilon_n(y_n)\big) && \text{(6a)}\\ &= \operatorname{clip}\big(\alpha(y_r) + \epsilon_n(\alpha(y_r))\big) && \text{(6b)}\\ &\overset{*}{=} \operatorname{clip}\big(\alpha(x_r + \epsilon_r(y_r)) + \epsilon_n(\alpha(y_r))\big) && \text{(6c)}\\ &\approx \operatorname{clip}\big(\alpha(x_r + \epsilon_r(x_r)) + \epsilon_n(\alpha(x_r))\big) && \text{(6d)}\\ &= \operatorname{clip}\big(\alpha(x_r) + \alpha_1\epsilon_r(x_r) + \epsilon_n(\alpha(x_r))\big). && \text{(6e)} \end{aligned}$$

The equality denoted with $*$ holds for non-clipped pixels in $x_r$, which are easily identified. The approximation defines the noise distributions in terms of the observed $x_r$ instead of the unknown intensities $y_r$, since $x_r$ is affected only little by noise. Exploiting that our capture protocol ensures that $\alpha_1$ and $\alpha_2$ are very close to 1 and 0, respectively, we can further approximate the scaled noise $\alpha_1\epsilon_r(x_r)$ as the noise of the linearly transformed image $\alpha(x_r)$:

$$\begin{aligned} &\alpha_1\epsilon_r(x_r) \sim \mathcal{N}\big(0, \alpha_1\sqrt{\beta_1^r x_r + \beta_2^r}\big) && \text{(7a)}\\ &\approx \mathcal{N}\Big(0, \sqrt{\beta_1^r(\alpha_1 x_r + \alpha_2) + \beta_2^r}\Big) \sim \epsilon_r(\alpha(x_r)). && \text{(7b)} \end{aligned}$$

For details see the supplementary material. We thus recover $\alpha$ from $x_n$ and $x_r$ by fitting the regression model

$$x_n \approx \operatorname{clip}\big(\alpha(x_r) + \epsilon_{r,n}(\alpha(x_r))\big), \tag{8}$$

where the parameters of the noise level function $\sigma_{r,n}$ of the compound noise $\epsilon_{r,n}$ are given by adding up the parameters $\boldsymbol{\beta}^r$ and $\boldsymbol{\beta}^n$ due to $\epsilon_r$ and $\epsilon_n$ being independent:

$$\begin{aligned} &\epsilon_{r,n}(x_r) \sim \mathcal{N}\big(0, \sigma_{r,n}(x_r)\big) && \text{(9a)}\\ \text{with} \quad &\sigma_{r,n}^2(x_r) = (\beta_1^r + \beta_1^n)x_r + (\beta_2^r + \beta_2^n). && \text{(9b)} \end{aligned}$$

Since the model defined in Eqs. (8) – (9b) accounts for both clipped observations as well as the heteroscedasticity of the noise, we call it *heteroscedastic Tobit regression*.

It generalizes basic Tobit regression [35], which only models clipped observations with homoscedastic noise. We can estimate the linear scaling parameters $\alpha_1, \alpha_2$ and the added noise variance parameters $\boldsymbol{\beta}^r + \boldsymbol{\beta}^n$ by maximizing the log-likelihood (see supplementary material). In Sec. 5 we demonstrate that faithful modeling of the image formation process with heteroscedastic Tobit regression is crucial for obtaining accurate estimates of $\alpha_1, \alpha_2$. Having recovered $\boldsymbol{\alpha}$, we use it to linearly transform the intensities of the low-ISO image to get an intermediate post-processed image

$$y'_p = \alpha(x_r) = \alpha_1 x_r + \alpha_2. \tag{10}$$

Figure 4c shows the difference image after the linear correction. The intensity-dependent bias is removed.

Since the noise parameters $\boldsymbol{\beta}^n, \boldsymbol{\beta}^r$ mainly depend on the ISO value and characteristics of the sensor [10], we record them in a controlled laboratory setting using our novel regression model, see Sec. 5.3. Hence, for post-processing our real dataset, we fix $\boldsymbol{\beta}^r$ as well as $\boldsymbol{\beta}^n$ and only recover $\boldsymbol{\alpha}$. In Sec. 5.3 we demonstrate the accuracy of our noise estimates by showing that they are in high agreement to those obtained from spatial averaging [13].

**Spatial misalignment.** We treat minuscule shifts of the camera as a global 2D translation that we wish to undo. While we have experimented with modern DFT-based subpixel alignment [16], we found that the classical Lucas-Kanade approach [23] works better. Despite its simplicity, it recovers the translation very well even under strong noise, see Sec. 5. Having estimated the translation parameters, we shift $y'_p$ using bilinear interpolation to obtain the next intermediate image $y''_p$. Note that interpolation results in some smoothing. This is not critical when translating $y'_p$, since it contains few high frequencies. We avoid interpolating $x_n$ as it contains many high frequencies due to the noise.

**Low-frequency residual correction.** As we can see in Fig. 4c, there remains a low-frequency pattern on the debiased residual image $R(y''_p)$. We account that to small changes in the ambient lighting. Also, when taking pictures under artificial illumination the rolling shutter effect will cause flickering of the light sources to appear as low-frequency banding artifacts. Thanks to the noise being zero-mean in the debiased domain we can estimate the low-frequency pattern LF by low-pass filtering of $R(y''_p)$:

$$\text{LF} = \text{smooth}\left(R(y''_p)\right) = \text{smooth}\left(\mathcal{A}(y''_p) - x_n\right). \tag{11}$$

The final post-processed image $y_p$ is obtained by subtracting the low-frequency pattern and inverting the debiasing step as

$$y_p = \mathcal{A}^{-1}\big(\mathcal{A}(y''_p) - \text{LF}\big). \tag{12}$$

We use a guided filter [17] with a large 40 pixel support for smoothing, which we found to remove structured residuals better than a Gaussian filter in case $\alpha_1$ is not estimated perfectly. Figure 4d shows the final debiased residual image $R(y_p)$ after the low-frequency correction. Now we can see a mostly zero-mean noise image as we expected, *c.f.* Eq. (3). While the filtering adds some structured residuals tightly localized along strong edges, the magnitude of the effect is small compared to the noise strength. Also, we see that the variance of the noise increases with the image intensity, as expected for heteroscedastic noise.

## 5. Experimental Validation

We now analyze and validate our approach on simulated data and demonstrate generalization to real image pairs.

### 5.1. Post-processing is effective

We first evaluate how accurately our post-processing can recover the transformation between the latent images $y_n$ and $y_r$. Therefore, we simulate the image formation process of the reference and noisy image (Fig. 3). Specifically, we use captured low-ISO images as latent images $y_n$ and generate the other latent image $y_r$ by sampling a random transformation consisting of a spatial translation, linear intensity changes, and an additive low-frequency pattern. From the latent images we generate the observations $x_n$ and $x_r$ by adding noise and clipping the image intensities. For realistic sampling of the transformations, we leverage statistics estimated on the captured dataset. Specifically, we sample random horizontal and vertical translations from $\mathcal{N}(0, 0.5)$. The slope and offset of the linear transformation are sampled from $\mathcal{N}(1, 0.05)$ and $\mathcal{N}(0, 0.0025)$, respectively. We generate the low-frequency pattern by sampling random Fourier coefficients weighted with a peaky Gaussian. We normalize the pattern in the spatial domain to have zero mean and a mean magnitude of 0.001. We finally simulate $x_n$ and $x_r$ by applying clipped Poisson-Gaussian noise to $y_n$ and $y_r$, respectively, *i.e.*

$$x_i \sim \text{clip}\big(\beta_1^i \mathcal{P}(y_i/\beta_1^i) + \mathcal{N}\big(0, \sqrt{\beta_2^i}\big)\big), i \in \{n, r\}. \tag{13}$$

To validate the estimation accuracy for a wide range of scenarios, we evaluate 11 different parameter settings for the noise, with $\beta_1^n$ ranging from $10^{-4}$ to $10^{-1}$ and $\beta_2^n$ ranging from $5 \cdot 10^{-8}$ to $10^{-2}$. This covers the range of noise parameters of the consumer cameras used for our dataset. For the reference image we use the noise level function of the Sony A7R at base ISO, *i.e.* $\beta_1^r \approx 2 \cdot 10^{-5}$, $\beta_2^r \approx 10^{-8}$. For each setting of noise parameters we run 100 trials in total.

We now study how well the proposed post-processing can undo the simulated transformations. First, we look at intensity scaling. Figures 5a and 5b show the root mean squared error (RMSE) of the estimated slope $\alpha_1$ and offset $\alpha_2$ of the linear intensity transformation. We compare our proposed Tobit regression model to several baselines: First, Tobit regression with homoscedastic noise [35], *i.e.* the noise strength is independent of image intensities. Next, homoscedastic and heteroscedastic linear least squares where the observations are assumed to be unclipped. Finally, we compare to the regression model of

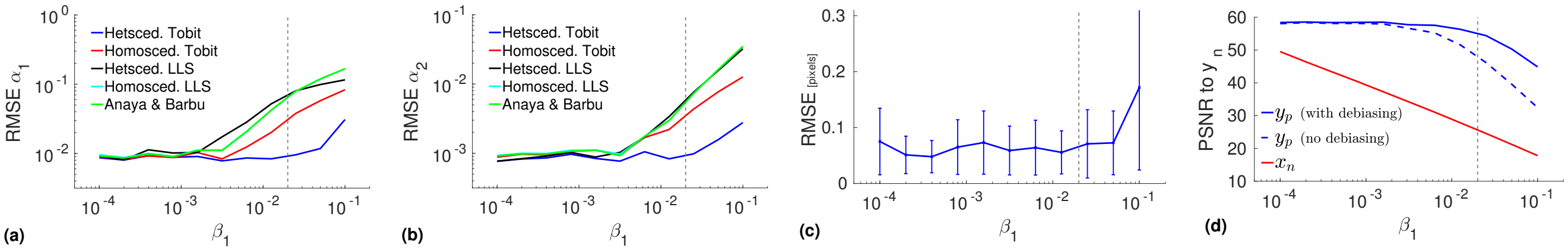

Figure 5. RMSE of recovering the slope (a) and offset (b) of simulated linear intensity scaling, and of recovering translation (c). PSNR to $y_n$ for the post-processed reference image $y_p$ and the noisy image $x_n$ (d). The $x$-axes show the strength of the intensity-dependent noise, with real values in our benchmark lying left of the gray dashed line.

[2], which models clipped observations while ignoring the intensity-dependence of the noise. We make two main observations: First, for low noise levels all methods perform equally well since the difficulty of the estimation problem is dominated by the other transformations, *i.e.* translation and low-frequency bias. Second, for medium to high noise levels our Tobit regression significantly outperforms all baselines including [2]. This shows the importance of modeling the clipped, heteroscedastic observation process faithfully.

Next we turn to alignment. Figure 5c shows the RMSE in pixels for recovering the simulated translation. As can be seen, the estimation error is robust to increasing levels of noise as it remains roughly constant over most of the range of noise settings. The error increases only for severe noise.

Finally, we evaluate the removal of low-frequency bias. Figure 5d shows the PSNR between the post-processed image $y_p$ and the latent image $y_n$. We compare our post-processing to a baseline (dashed) that omits the debiasing step of Eqs. (11) – (12). Especially for high noise levels, the PSNR of the baseline is significantly lower, emphasizing that filtering in the debiased domain is important. We note that the PSNR of $y_p$ reduces with higher noise levels, since the filtering step is not perfect and thus leaks low frequencies of the noise into the post-processed image. This is not critical, however, since the gap of the PSNR of the noisy image $x_n$ to that of the latent image is still large enough to accurately measure state-of-the-art denoising performance.

## 5.2. Quality of ground truth

We now demonstrate that our post-processing pipeline provides accurate denoising ground truth on our real-world dataset by considering statistics of the debiased residual images. We have already seen that the *ground truth residual* $R(y_n)$ has mean zero given $y_n$. It follows that $R(y_n)$ and $\mathcal{A}(y_n)$ are linearly uncorrelated (see supplemental material). Furthermore, when assuming pixel-wise independent noise, $R(y_n)$ has zero auto-correlation. We thus expect the post-processing residual $R(y_p)$ to have small linear correlation to $\mathcal{A}(y_p)$ as well as small auto-correlation. Moreover, we expect $R(y_p)$ to have a slightly higher variance than $R(y_n)$, since $R(y_p)$ also includes the small amount of noise that affects $x_r$.

We evaluate the three statistics of $R(y_p)$ on our real-world dataset as well as on simulated data. To make the simulation as realistic as possible, for each image we use the parameters for translation and intensity scaling that were obtained by running post-processing on the real data and use the corresponding noise level functions. Table 2 shows the mean absolute linear correlation coefficient $\text{Corr}(R(y_p), \mathcal{A}(y_p))$, the mean absolute auto-correlation $\text{Auto-Corr}(R(y_p))$, and the geometric mean of the variance $\text{Var}(R(y_p))$. We observe a significant linear correlation when not applying any post-processing to $x_r$. Our full post-processing pipeline almost completely removes the correlation as expected from theoretical considerations, highlighting its need for obtaining a database with realistic image noise. Note that just applying the high-pass filter on the residual image (5th row) still leaves a significant linear correlation, and that the combination of all three post-processing steps improves upon using any two post-processing steps. The same holds for auto-correlation, where our post-processing successfully obtains a residual image with low auto-correlation, indicating that the pixels in the noisy residual image are not highly spatially correlated. It is important to note that when only intensity scaling is applied, the auto-correlation is $5\times$ as high on real data. Since this is the only form of post-processing in the RENOIR dataset [2], we can conclude that our approach leads to a much more realistic image noise dataset.

Turning to variance, we see that the variance of the post-processing residual $R(y_p)$ is significantly closer to that of the ground truth residual $R(y_n)$ when all steps are carried out. The remaining gap to the ground truth residual can be explained as follows: The post-processed residual is affected by noise in $x_r$ *and* $x_n$, while the ground truth residual is affected only by the noise in $x_n$. We thus also computed the variance of the ground truth residual $R(y_n)$ for a second setting, where we sample the noise of $x_n$ from the compound noise $\epsilon_{r,n}$ (Eq. 9a) instead of $\epsilon_n$. Then the difference in variance between post-processed and ground truth residual almost vanishes, and the relative variance error decreases by an order of magnitude compared to no post-processing. This demonstrates that our post-processing removes the global effects on the residual image while accu-rately preserving the noise characteristics.

| Intensity scaling | Alignment | LF correction | $\text{Corr}(R(y_p), \mathcal{A}(y_p))$ | | $\text{Auto-Corr}(R(y_p))$ | | $\text{Var}(R(y_p))$ [$*10^{-3}$] | | $\text{PSNR}(y_p, y_n)$ |
|---|---|---|---|---|---|---|---|---|---|
| | | | synth | real | synth | real | synth | real | synth [dB] |
| | | | 0.2144 | 0.1874 | 0.1407 | 0.1270 | 0.1921 | 0.1815 | 43.14 |
| ✓ | | | 0.0305 | 0.0318 | 0.0923 | 0.0843 | 0.1752 | 0.1690 | 46.45 |
| | ✓ | | 0.2093 | 0.1892 | 0.0958 | 0.1024 | 0.1482 | 0.1583 | 46.37 |
| ✓ | ✓ | | 0.0418 | 0.0474 | 0.0478 | 0.0560 | 0.1387 | 0.1473 | 51.18 |
| | | ✓ | 0.0170 | 0.0175 | 0.0615 | 0.0581 | 0.1659 | 0.1626 | 47.31 |
| ✓ | | ✓ | 0.0078 | 0.0067 | 0.0610 | 0.0559 | 0.1656 | 0.1621 | 47.56 |
| | ✓ | ✓ | 0.0118 | 0.0140 | 0.0066 | 0.0198 | 0.1313 | 0.1389 | 53.13 |
| ✓ | ✓ | ✓ | 0.0029 | 0.0051 | 0.0067 | 0.0173 | 0.1314 | 0.1385 | 53.71 |

Table 2. Statistics of the residual noise image for different combinations of post-processing steps on both synthetic and real data. For reference: $\text{Var}(R(y_n)) = 0.1222 \cdot 10^{-3}$, respectively $\text{Var}(R(y_n)) = 0.1356 \cdot 10^{-3}$ when sampling noise for $x_n$ from $\epsilon_{r,n}$ instead of $\epsilon_n$.

Importantly, the three test statistics obtained from synthetic experiments differ only marginally from those evaluated on the real captured images, showing that the modeled transformation process consisting of translations, intensity scaling, and an additive low-frequency pattern accurately describes the real transformation between $y_n$ and $y_r$.

Finally, Table 2 also shows the PSNR between $y_n$ and $y_p$ on simulated data. We see that our full post-processing achieves the highest PSNR of 53.7 dB. This is significantly more than what state-of-the-art denoising algorithms can currently achieve (Sec. 6), leaving enough room for measuring future improvements in terms of PSNR.

### 5.3. Recording of noise parameters

We calibrate the noise parameters $\boldsymbol{\beta}^r$ and $\boldsymbol{\beta}^n$ on controlled test scenes of a color checker. To estimate $\boldsymbol{\beta}^r$, we first run Tobit regression on pairs of images, both taken at base ISO, which yields an estimate of $2\boldsymbol{\beta}^r$ (Eq. 9b). We subsequently recover $\boldsymbol{\beta}^n$ for all other ISO values by estimating $\boldsymbol{\beta}^r + \boldsymbol{\beta}^n$ on a low/high-ISO image pair and afterwards subtracting $\boldsymbol{\beta}^r$. To assess the accuracy of our estimates we compare them to those obtained from the individual images using the spatial averaging method of [13], which is designed to work highly accurately on images with piecewise constant intensities. We assess the agreement of both methods with the normalized RMSE $\Phi$ proposed in [27]. It measures the relative error of standard deviations, averaged over pixel intensities. Specifically, we use the symmetric extension

$$\tilde{\Phi}(\boldsymbol{\beta}, \hat{\boldsymbol{\beta}}) = \tfrac{1}{2}\big(\Phi(\boldsymbol{\beta}, \hat{\boldsymbol{\beta}}) + \Phi(\hat{\boldsymbol{\beta}}, \boldsymbol{\beta})\big). \tag{14}$$

The mean error between Tobit regression and [13] is 0.003, *i.e.* standard deviations from both methods disagree only marginally by $0.3\%$ on average. We conclude that Tobit regression produces accurate noise estimates on real data. But unlike spatial averaging methods, it generalizes to arbitrary scenes without large homogeneous areas.

We now justify using calibrated noise parameters for post-processing by showing that the noise parameters mainly depend on ISO value and camera, but not on absolute exposure time. For fixed combinations of ISO and camera, we estimate $\boldsymbol{\beta}^r$ and $\boldsymbol{\beta}^n$ across a range of exposure times of the image pairs. The average error $\tilde{\Phi}$ between those noise estimates is only $0.5\%$, showing that they are stable w.r.t. overall exposure times.

## 6. Benchmark

The proposed DND benchmark for denoising algorithms consists of 50 scenes selected from our captured images. We chose images that look like typical photographs, but also included images with interesting structures that we believe to be challenging for the algorithms tested. A subset of the test images is shown in Fig. 2. Table 1 lists the number of scenes per camera included in the benchmark dataset.

For the task of (non-blind) denoising, we compare the performance of Weighted Nuclear Norm Minimization (WNNM) [15], K-SVD [1], Expected Patch Log Likelihood (EPLL) [39], Field of Experts (FoE) [32] with the filters of [14], Nonlocally Centralized Sparse Representations (NCSR) [8], and BM3D [7]. Moreover, we benchmark two discriminative, "deep" methods: A multilayer network (MLP) [5] and Trainable Nonlinear Reactive Diffusion (TNRD) [6]. For MLP, we use available trained models for Gaussian noise with $\sigma \in \{10, 25, 35\}$. TNRD is trained on 400 separate images [6] using code from the authors' web page. We train 10 models with different Gaussian noise standard deviations, evenly distributed in log-space from $0.0001$ to $0.1$, thus covering a reasonable range of noise levels observed on our real-world dataset.

We apply all algorithms to the noisy images in three different spaces. First, we use the space of linear raw intensities. Since the tested methods are mostly geared toward Gaussian denoising, we apply a variance stabilizing transformation (VST) prior to denoising as a second setting. This has the effect of approximately Gaussianizing the noise distribution. After retrieving the denoising result, we convert it back to linear raw space by applying an inverse VST. Specifically, we use the generalized Anscombe transform

| Applied on | Evaluated on | WNNM | KSVD | EPLL | FoE | NCSR | BM3D | MLP | TNRD |
|---|---|---|---|---|---|---|---|---|---|
| RAW | RAW | 46.29 | 45.53 | 46.34 | 45.77 | 42.86 | **46.63** | 42.70 | 44.98 |
| RAW | sRGB | 37.64 | 36.69 | 37.27 | 36.09 | 30.97 | **37.86** | 33.74 | 35.69 |
| RAW+VST | RAW | 47.10 | 46.86 | 46.85 | 44.11 | 47.06 | **47.14** | 45.70 | 45.69 |
| RAW+VST | sRGB | **37.97** | 37.72 | 37.55 | 35.97 | 37.85 | 37.95 | 36.83 | 36.22 |
| sRGB | sRGB | 34.44 | **36.55** | 33.51 | 34.49 | 33.81 | 34.61 | 34.14 | 29.92 |

Table 3. Mean PSNR (in dB) of the denoising methods tested on our DND benchmark. We apply denoising either on linear raw intensities, after a variance stabilizing transformation (VST), or after conversion to the sRGB space. Likewise, we evaluate the result either in linear raw space or in sRGB space. The noisy images have a PSNR of 39.39 dB (linear raw) and 29.98 dB (sRGB).

[34] and the closed-form approximation to its exact unbiased inverse [26]. We parametrize the transformation with the noise-level functions obtained from the color-checker data (Sec. 5.3). In a third setting, we use available EXIF data to simulate the main steps of the camera processing pipeline [20] that converts linear raw intensities to sRGB intensities. After white-balancing, we demosaic the image by linear interpolation. Finally, we convert from the camera internal color space to sRGB and apply gamma correction.

Since many of the benchmarked algorithms are too slow to be applied to megapixel-sized images, we crop 20 bounding boxes of $512 \times 512$ pixels from each image in the dataset, yielding 1000 test crops in total. They overlap at most 10% and do not contain pixels that were annotated as changing between the two exposures. We provide the algorithms with an estimate of the global noise standard deviation $\bar{\sigma}$ by computing the standard deviation of the residual noise image $R(y_p)$ on each crop. As the different color channels usually look quite distinct, we denoise each channel separately. For TNRD and MLP we choose the model whose $\sigma$ for training is closest to the ground truth $\bar{\sigma}$. For FoE and EPLL we use a heteroscedastic Gaussian data term when denoising raw pixel intensities and a homoscedastic Gaussian data term in the other cases. For evaluation, we compare the denoised result to the post-processed reference image $y_p$ either in linear raw space or in the sRGB space.

Table 3 shows the PSNR values, averaged over all crops and color channels (SSIM values [36] are available in the supplemental material). We make several interesting observations. As we can see, BM3D is overall the best performing method followed by WNNM. The other methods perform worse. The general tendency also holds across noise levels. This is quite surprising as the by now classic BM3D approach was previously considered to have been outperformed by the other approaches; our realistic noise dataset shows that this is not the case. The discriminative methods fall short, which suggests that they generalize poorly to noise distributions that were not used during training. The generative FoE model performs surprisingly competitive in linear raw space, but is the only baseline that performs worse after VST. This suggests that FoE benefits from the more realistic likelihood in linear raw space.

Furthermore, we see that denoising sRGB images yields significantly worse results than applying denoising algorithms in raw space, since the noise distribution in sRGB space is spatio-chromatically correlated [30]. Another observation is that the amount of noise in our realistic dataset is lower than what is often used in the scientific literature for evaluating denoising algorithms using synthetic noise. The mean PSNR of the noisy images in raw space is 39.38 dB, which would correspond to a mean noise standard deviation of $\sigma \approx 2.74$ for images with intensities in $[0, 255]$. For comparison, most denoising algorithms are evaluated with noise standard deviations of at least $\sigma = 10$, which we believe to be mostly a historical artefact. Apparently, it was never really questioned whether they are still appropriate.

## 7. Conclusion

To benchmark denoising algorithms on real photographs, we introduced an acquisition procedure based on pairs of images of the same scene, captured with different analog gains and exposure time. While in theory the per-pixel mean intensity should stay constant, in practice we encountered residual errors. To derive ground-truth data, we proposed and evaluated a procedure for handling residual errors stemming from inaccurate gain and exposure time changes, relying on a novel heteroscedastic Tobit regression model. We also correct for lighting changes in a transformed space, as well as spatial misalignments. Our experiments showed the efficacy of this post-processing on simulated data, as well as its necessity on real photographs. We will make our novel ground-truth dataset of real photographs publicly available as a benchmark. We used it for evaluating various denoising algorithms and observed that BM3D continues to outperform recent denoising methods on real photographs, which is in contrast to findings on previously considered synthetic settings. More generally, our analysis revealed that the common scientific practice for evaluating denoising techniques has rather limited relevance for realistic settings.

**Acknowledgments:** The research leading to these results has received funding from the European Research Council under the European Union's Seventh Framework Programme (FP/2007-2013)/ERC Grant agreement No. 307942, as well as from the EU FP7 project "Harvest4D" (No. 323567).

# Benchmarking Denoising Algorithms with Real Photographs
## – Supplemental Material –

Tobias Plötz Stefan Roth
Department of Computer Science, TU Darmstadt

**Preface.** In this supplemental material we give a proof for $\mathcal{A}(y_n)$ and $R(y_n)$ being linearly uncorrelated. We, furthermore, give additional details on our novel heteroscedastic Tobit regression model (derivation, log-likelihood and its gradient) and highlight the importance of considering clipping of the noisy observations. Finally, we show additional results from our denoising benchmark.

## A. Linear Correlation of $\mathcal{A}(y_n)$ and $R(y_n)$

**Proposition 1.** *The debiased image $\mathcal{A}(y_n)$ and the debiased residual image $R(y_n)$ are linearly uncorrelated.*

*Proof.* First, we note that the expectation of $R(y_n)$ given $\mathcal{A}(y_n)$ is zero

$$\begin{aligned}
&\mathbb{E}\left[R(y_n) \mid \mathcal{A}(y_n)\right] \\
&\quad= \mathbb{E}\left[\mathcal{A}(y_n) - x_n \mid \mathcal{A}(y_n)\right] && \text{(15a)}\\
&\quad= \mathbb{E}\left[\mathcal{A}(y_n) \mid \mathcal{A}(y_n)\right] - \mathbb{E}\left[x_n \mid \mathcal{A}(y_n)\right] && \text{(15b)}\\
&\quad= \mathcal{A}(y_n) - \mathbb{E}\left[x_n \mid y_n\right] && \text{(15c)}\\
&\quad= 0, && \text{(15d)}
\end{aligned}$$

where the third equality follows from the fact that $\mathcal{A}(\cdot)$ is invertible [11]. Next, we observe that for two random variables $X$ and $Y$, the expectation of $X$ is zero if $\mathbb{E}\left[X \mid y = Y\right] = 0$ for all $y$:

$$\mathbb{E}\left[X\right] = \mathbb{E}_Y\left[\mathbb{E}_{X|Y}\left[X\right]\right] = \mathbb{E}_Y\left[0\right] = 0. \tag{16}$$

We now show that two random variables $X$ and $Y$ have zero covariance if $\mathbb{E}\left[X \mid y = Y\right] = 0$ for all $y$:

$$\begin{aligned}
\mathrm{Cov}(X, Y) &= \mathbb{E}\big[(X - \mathbb{E}X)(Y - \mathbb{E}Y)\big] && (17)\\
&= \mathbb{E}\big[X(Y - \mathbb{E}Y)\big] && (18)\\
&= \mathbb{E}\left[XY\right] - \mathbb{E}X \cdot \mathbb{E}Y && (19)\\
&= \mathbb{E}_Y\left[\mathbb{E}_{X|Y}\left[XY\right]\right] && (20)\\
&= \mathbb{E}_Y\left[Y \cdot \mathbb{E}_{X|Y}\left[X\right]\right] && (21)\\
&= \mathbb{E}_Y\left[Y \cdot 0\right] && (22)\\
&= 0. && (23)
\end{aligned}$$

From the definition of the linear correlation coefficient it follows that zero covariance between two random variables implies that they are linearly uncorrelated. □

## B. Heteroscedastic Tobit Regression

We now derive the log-likelihood and its gradient of the proposed heteroscedastic Tobit regression model (Eqs. 8–9b in the paper). Moreover, we detail the approximation of the noise term of Eqs. (7a) – (7b) of the main paper. For clarity, we denote $\alpha(x_r) = \boldsymbol{\alpha}^\top \mathbf{x} \doteq \tilde{x}$, where $\mathbf{x} = [x_r, 1]^\top$.

### B.1. Log-likelihood

Before deriving the log-likelihood of Eq. (8), let us first look at the theoretical case of unclipped intensities $x'_n$ in the high-ISO image:

$$x'_n = \tilde{x} + \epsilon_{r,n}(\tilde{x}). \tag{24}$$

Following Eq. (9a) of the main paper, the conditional distribution of $x'_n$ given the intensities in $\tilde{x}$ is given as a heteroscedastic Gaussian:

$$p(x'_n \mid x_r) = \mathcal{N}\big(x'_n \mid \tilde{x},\ \sigma_{r,n}(\tilde{x})\big). \tag{25}$$

We now consider the clipped noisy signal $x_n$. To derive its conditional distribution in case that $x_n$ is clipped, we replace the Gaussian PDF with Dirac deltas weighted by the probability mass of all possible values $x'_n$ that would be clipped to $x_n$. Hence, the conditional distribution is given by a case distinction on whether $x_n$ is unclipped, clipped from below, or from above, respectively. Precisely, we can write

$$\mathcal{T}(x_n \mid x_r) = \begin{cases} \mathcal{N}\big(x_n \mid \tilde{x}, \sigma_{r,n}(\tilde{x})\big), & \text{if } 0 < x_n < 1 \\ \delta(x_n) \cdot \int_{-\infty}^{0} \mathcal{N}\big(x'_n \mid \tilde{x}, \sigma_{r,n}(\tilde{x})\big)\, \mathrm{d}x'_n, & \text{if } x_n \leq 0 \\ \delta(1 - x_n) \cdot \int_{1}^{\infty} \mathcal{N}\big(x'_n \mid \tilde{x}, \sigma_{r,n}(\tilde{x})\big)\, \mathrm{d}x'_n, & \text{if } x_n \geq 1. \end{cases} \tag{26}$$

It is easy to check that $\mathcal{T}(x_n \mid x_r)$ indeed is a valid probability distribution. Obviously, $\mathcal{T}(x_n \mid x_r) \geq 0$ and

$$\int_{\mathbb{R}} \mathcal{T}(x_n \mid x_r)\, \mathrm{d}x_n = \int_{-\infty}^{0} \mathcal{T}(x_n \mid x_r)\, \mathrm{d}x_n + \int_{0}^{1} \mathcal{T}(x_n \mid x_r)\, \mathrm{d}x_n + \int_{1}^{\infty} \mathcal{T}(x_n \mid x_r)\, \mathrm{d}x_n \tag{27a}$$

$$= \int_{-\infty}^{0} \mathcal{N}\big(x_n' \mid \tilde{x}, \sigma_{r,n}(\tilde{x})\big)\, \mathrm{d}x_n' + \int_{0}^{1} \mathcal{N}\big(x_n \mid \tilde{x}, \sigma_{r,n}(\tilde{x})\big)\, \mathrm{d}x_n + \int_{1}^{\infty} \mathcal{N}\big(x_n' \mid \tilde{x}, \sigma_{r,n}(\tilde{x})\big)\, \mathrm{d}x_n' \tag{27b}$$

$$= 1. \tag{27c}$$

By denoting the cumulative distribution function of a standard normal distribution as $\Psi(z) = \int_{-\infty}^{z} \mathcal{N}(z' \mid 0, 1)\, \mathrm{d}z'$ and by noting that $\Psi\big(\frac{z-\mu}{\sigma}\big) = \int_{-\infty}^{z} \mathcal{N}(z' \mid \mu, \sigma)\, \mathrm{d}z'$, we can write the log-likelihood of $\mathcal{T}(x_n \mid x_r)$ up to constants as

$$\log \mathcal{T}(x_n \mid x_r) = \begin{cases} -\log \sigma_{r,n}(\tilde{x}) - \frac{(x_n - \tilde{x})^2}{2\sigma_{r,n}(\tilde{x})^2}, & \text{if } 0 < x_n < 1 \\ \delta(x_n) \cdot \log \Psi\left(\frac{-\tilde{x}}{\sigma_{r,n}(\tilde{x})}\right), & \text{if } x_n \leq 0 \\ \delta(1 - x_n) \cdot \log\left(1 - \Psi\left(\frac{1-\tilde{x}}{\sigma_{r,n}(\tilde{x})}\right)\right), & \text{if } x_n \geq 1. \end{cases} \tag{28}$$

For constant $\sigma_{r,n}(\tilde{x}) = \sigma_{r,n}$ (*i.e.*, stationary noise) this is the log-likelihood of Tobit regression with clipping at 0 from below and at 1 from above [35]. In our special case, we use a non-constant link function for the standard deviation, *i.e.*

$$\sigma_{r,n}(\tilde{x}) = \sqrt{\beta_1^{r,n}\tilde{x} + \beta_2^{r,n}} \tag{29a}$$

$$= \sqrt{(\beta_1^r + \beta_1^n)\tilde{x} + \beta_2^r + \beta_2^n} \tag{29b}$$

in order to define our heteroscedastic Tobit regression model.

To estimate its parameters, we minimize the negative log-likelihood of all data points

$$(\hat{\boldsymbol{\alpha}}, \hat{\boldsymbol{\beta}}) = \operatorname*{arg\,min}_{\boldsymbol{\alpha}, \boldsymbol{\beta}^{r,n}} \sum_i -\log \mathcal{T}\big(x_n^{(i)} \mid x_r^{(i)}\big). \tag{30}$$

## B.2. Log-likelihood Gradient

It is useful to first derive the partial derivatives of terms of the form ${(c-\tilde{x})}/{\sigma_{r,n}(\tilde{x})}$ for some constant $c$ w.r.t. all variables. The partial derivatives can be shown to be given as:

$$\frac{\partial\, {(c-\tilde{x})}/{\sigma_{r,n}(\tilde{x})}}{\partial \beta_1^{r,n}} = -\frac{1}{2}\big(c - \boldsymbol{\alpha}^\top \mathbf{x}\big) \cdot \big(\beta_1^{r,n}\boldsymbol{\alpha}^\top \mathbf{x} + \beta_2^{r,n}\big)^{-3/2} \cdot \boldsymbol{\alpha}^\top \mathbf{x} \tag{31a}$$

$$\frac{\partial\, {(c-\tilde{x})}/{\sigma_{r,n}(\tilde{x})}}{\partial \beta_2^{r,n}} = -\frac{1}{2}\big(c - \boldsymbol{\alpha}^\top \mathbf{x}\big) \cdot \big(\beta_1^{r,n}\boldsymbol{\alpha}^\top \mathbf{x} + \beta_2^{r,n}\big)^{-3/2} \tag{31b}$$

$$\frac{\partial\, {(c-\tilde{x})}/{\sigma_{r,n}(\tilde{x})}}{\partial \boldsymbol{\alpha}} = -\mathbf{x} \cdot \big(\beta_1^{r,n}\boldsymbol{\alpha}^\top \mathbf{x} + \beta_2^{r,n}\big)^{-1/2} - \frac{1}{2}\big(c - \boldsymbol{\alpha}^\top \mathbf{x}\big) \cdot \big(\beta_1^{r,n}\boldsymbol{\alpha}^\top \mathbf{x} + \beta_2^{r,n}\big)^{-3/2} \cdot \beta_1^{r,n}\mathbf{x}. \tag{31c}$$

That allows to derive the partial derivatives for all three cases of the log-likelihood function. For the first case they are given as

$$\frac{\partial \log \mathcal{N}\big(x_n \mid \tilde{x}, \sigma_{r,n}(\tilde{x})\big)}{\partial \beta_1^{r,n}} = -\frac{1}{2}\big(\beta_1^{r,n}\boldsymbol{\alpha}^\top \mathbf{x} + \beta_2^{r,n}\big)^{-1} \cdot \boldsymbol{\alpha}^\top \mathbf{x} - \frac{x_n - \boldsymbol{\alpha}^\top \mathbf{x}}{\sqrt{\beta_1^{r,n}\boldsymbol{\alpha}^\top \mathbf{x} + \beta_2^{r,n}}} \cdot \frac{\partial\, {(x_n-\tilde{x})}/{\sigma_{r,n}(\tilde{x})}}{\partial \beta_1^{r,n}} \tag{32a}$$

$$\frac{\partial \log \mathcal{N}\big(x_n \mid \tilde{x}, \sigma_{r,n}(\tilde{x})\big)}{\partial \beta_2^{r,n}} = -\frac{1}{2}\big(\beta_1^{r,n}\boldsymbol{\alpha}^\top \mathbf{x} + \beta_2^{r,n}\big)^{-1} - \frac{x_n - \boldsymbol{\alpha}^\top \mathbf{x}}{\sqrt{\beta_1^{r,n}\boldsymbol{\alpha}^\top \mathbf{x} + \beta_2^{r,n}}} \cdot \frac{\partial\, {(x_n-\tilde{x})}/{\sigma_{r,n}(\tilde{x})}}{\partial \beta_2^{r,n}} \tag{32b}$$

$$\frac{\partial \log \mathcal{N}\big(x_n \mid \tilde{x}, \sigma_{r,n}(\tilde{x})\big)}{\partial \boldsymbol{\alpha}} = -\frac{1}{2}\big(\beta_1^{r,n}\boldsymbol{\alpha}^\top \mathbf{x} + \beta_2^{r,n}\big)^{-1} \cdot \beta_1^{r,n}\mathbf{x} - \frac{x_n - \boldsymbol{\alpha}^\top \mathbf{x}}{\sqrt{\beta_1^{r,n}\boldsymbol{\alpha}^\top \mathbf{x} + \beta_2^{r,n}}} \cdot \frac{\partial\, {(x_n-\tilde{x})}/{\sigma_{r,n}(\tilde{x})}}{\partial \boldsymbol{\alpha}}. \tag{32c}$$

To compute the last term of each equation we employ Eqs. (31a) – (31c). For the second case the gradient of the

| Applied on | Evaluated on | WNNM | KSVD | EPLL | NCSR | BM3D | MLP | TNRD | FoE |
|---|---|---|---|---|---|---|---|---|---|
| RAW | RAW | 0.971 | 0.968 | 0.968 | 0.853 | **0.972** | 0.939 | 0.963 | 0.967 |
| RAW | sRGB | **0.933** | 0.919 | 0.931 | 0.713 | **0.933** | 0.886 | 0.894 | 0.907 |
| RAW+VST | RAW | **0.974** | 0.972 | 0.973 | 0.969 | **0.974** | 0.963 | 0.961 | 0.955 |
| RAW+VST | sRGB | **0.935** | 0.931 | 0.927 | 0.924 | 0.932 | 0.916 | 0.892 | 0.914 |
| sRGB | sRGB | 0.866 | **0.900** | 0.829 | 0.834 | 0.855 | 0.838 | 0.708 | 0.887 |

Table 4. Mean SSIM [36] of the denoising methods tested on our benchmark dataset. We apply denoising either on linear raw intensities, after a variance stabilizing transformation (VST), or after conversion to the sRGB space. Likewise, we evaluate the result either in linear raw space or in sRGB space. The noisy images have a SSIM of 0.863 (linear raw) and 0.710 (sRGB).

log-likelihood is given by

$$\frac{\partial \log \Psi\left(\frac{-\tilde{x}}{\sigma_{r,n}(\tilde{x})}\right)}{\partial \beta_1^{r,n}} = \frac{\mathcal{N}\left(0 \mid \tilde{x}, \sigma_{r,n}(\tilde{x})\right)}{\Psi\left(\frac{-\tilde{x}}{\sigma_{r,n}(\tilde{x})}\right)} \cdot \frac{\partial (-\tilde{x})/\sigma_{r,n}(\tilde{x})}{\partial \beta_1^{r,n}} \tag{33a}$$

$$\frac{\partial \log \Psi\left(\frac{-\tilde{x}}{\sigma_{r,n}(\tilde{x})}\right)}{\partial \beta_2^{r,n}} = \frac{\mathcal{N}\left(0 \mid \tilde{x}, \sigma_{r,n}(\tilde{x})\right)}{\Psi\left(\frac{-\tilde{x}}{\sigma_{r,n}(\tilde{x})}\right)} \cdot \frac{\partial (-\tilde{x})/\sigma_{r,n}(\tilde{x})}{\partial \beta_2^{r,n}} \tag{33b}$$

$$\frac{\partial \log \Psi\left(\frac{-\tilde{x}}{\sigma_{r,n}(\tilde{x})}\right)}{\partial \boldsymbol{\alpha}} = \frac{\mathcal{N}\left(0 \mid \tilde{x}, \sigma_{r,n}(\tilde{x})\right)}{\Psi\left(\frac{-\tilde{x}}{\sigma_{r,n}(\tilde{x})}\right)} \cdot \frac{\partial (-\tilde{x})/\sigma_{r,n}(\tilde{x})}{\partial \boldsymbol{\alpha}}, \tag{33c}$$

again employing Eqs. (31a) – (31c). The third case works analogously. In practice, we optimize for $\boldsymbol{\beta}' = \log \boldsymbol{\beta}^{r,n}$ to ensure that $\boldsymbol{\beta}^{r,n}$ is positive. Furthermore, we exclude pixels near image edges [13] from the regression and truncate the log-likelihood to be robust to outliers, *i.e.* we set the gradients to zero for pixels with $\log \mathcal{T}(x_n^i \mid x_r^i) < -10$.

When estimating the $\boldsymbol{\alpha}$ parameter for the image pairs in our dataset, we use previously recorded noise parameters $\boldsymbol{\beta}$. These were obtained from running our full Tobit regression on controlled images showing a color checker, see Fig. 6.

## B.3. Approximation of Noise Term

Here, we quantify the error that is induced by approximating the noise term in Eqs. (7a) – (7b) of the main paper. Specifically, we approximate the variance of the Gaussian noise by

$$\alpha_1^2 \left(\beta_1^r x_r + \beta_2^r\right) \approx \beta_1^r (\alpha_1 x_r + \alpha_2) + \beta_2^r. \tag{34}$$

Obviously, the left-hand side would converge to the right-hand side as $\alpha_1 \to 1$ and $\alpha_2 \to 0$, if the ISO value and exposure time were set with perfect accuracy. In practice, however, this is not the case. We now evaluate the practical impact of our approximation. With denoting $\beta(x_r) = \beta_1 x_r + \beta_2$, let

$$\sigma(x_r) = \sqrt{\alpha_1^2 \beta^r(x_r) + \beta^n(\alpha(x_r))} \tag{35}$$

be the true noise level function and

$$\hat{\sigma}(x_r) = \sqrt{\beta^{r,n}(\alpha(x_r))} \tag{36}$$

be the approximated noise level function. We compute the normalized root mean squared error $\Phi$ [27] between the true and the approximated noise level function, assuming a uniform distribution over pixel intensities

$$\Phi(\sigma, \hat{\sigma}) = \int_0^1 \frac{\left(\sigma(x_r) - \hat{\sigma}(x_r)\right)^2}{\sigma(x_r)} \, \mathrm{d}x_r. \tag{37}$$

The average normalized RMSE on our dataset is $1.4 \cdot 10^{-4}$, meaning that on average approximating the noise standard deviation introduces a relative error of $0.014\%$. This is insignificant compared to the overall estimation accuracy (see Sec. 5 of the paper).

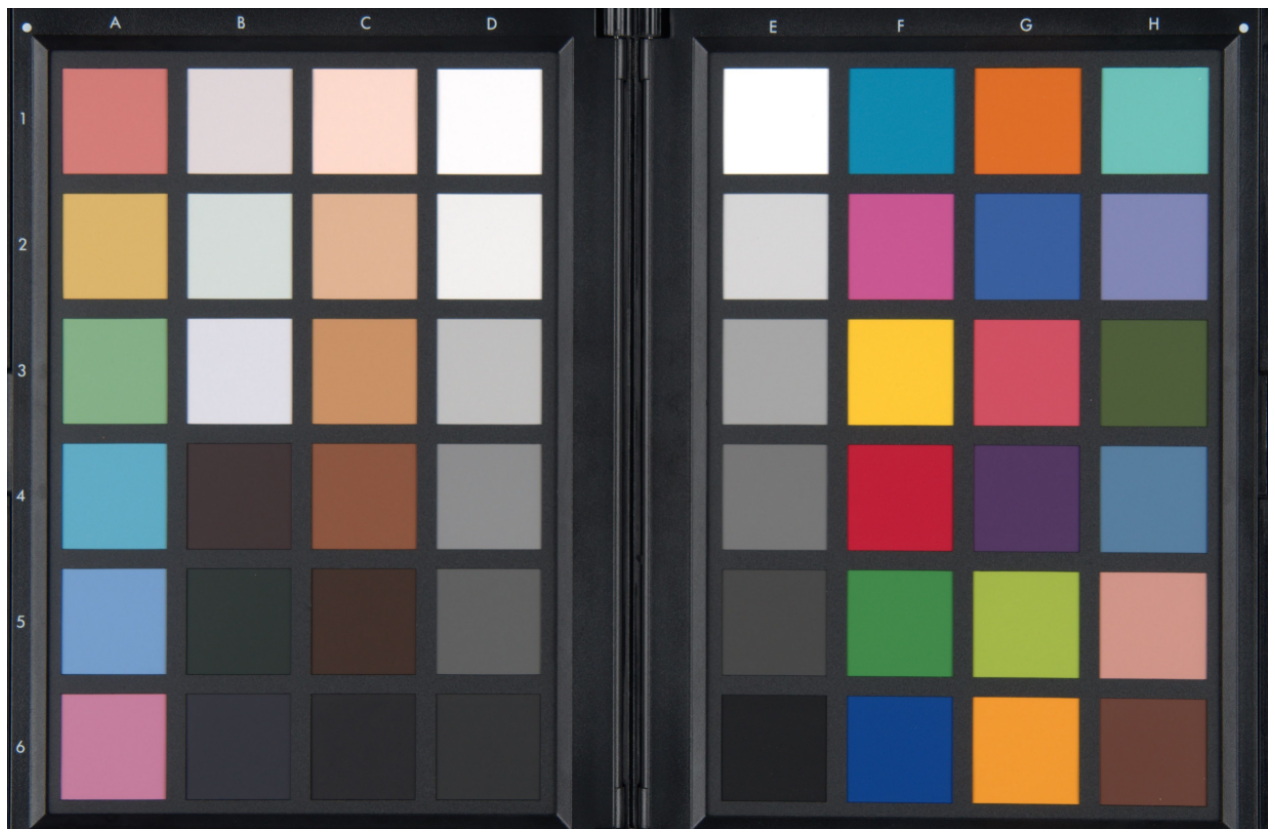

Figure 6. Test scene used for noise parameter calibration.

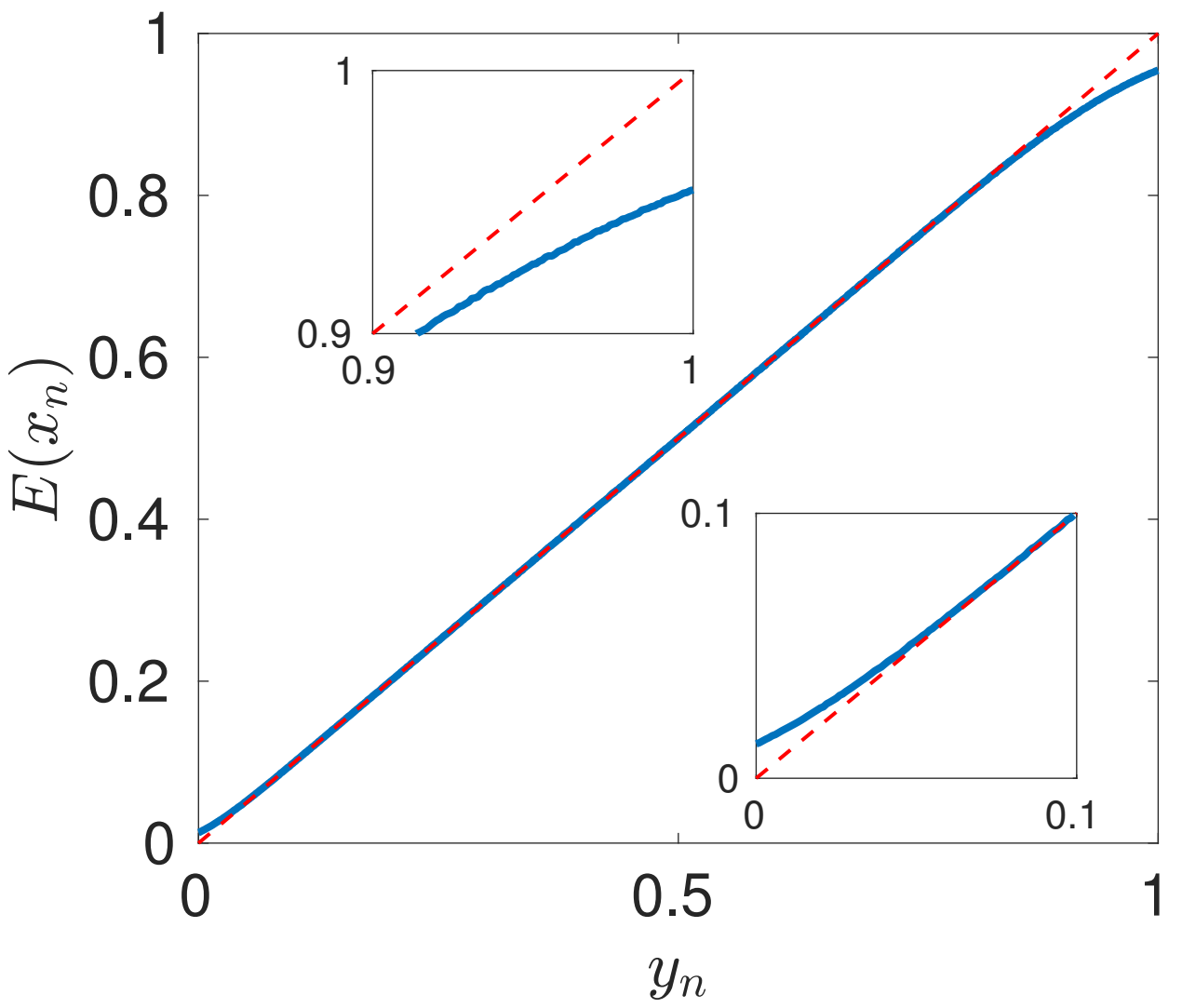

Figure 7. Noise-free intensities (red dashed line) *vs*. mean of clipped noisy intensities (blue solid line).

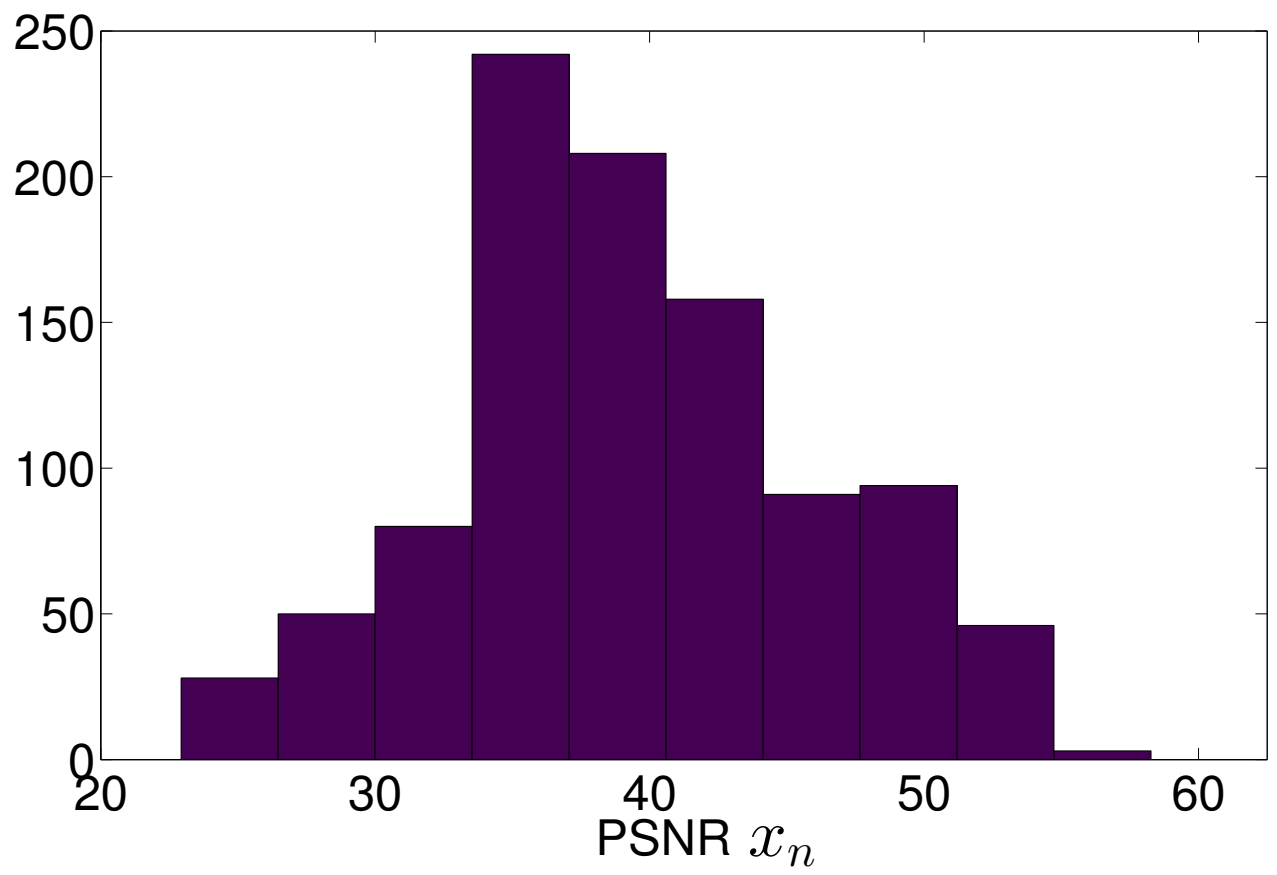

Figure 8. Histogram of PSNR values (in dB) of the crops of the noisy test images.

## C. Bias from Clipping

Figure 7 plots the noise-free image intensities $y_n$ against the average of clipped noisy observations $x_n$ for the noise level function of the Nexus 6P at ISO 6400. We can see that the mean of the clipped observations strongly deviates from the true noise-free intensities near the clipping boundaries, also see [11]. Due to this bias introduced by clipping the signal, we can not recover the noise free signal by simply averaging noisy observations spatially or temporally. Hence, we perform the smoothing operation of our low-frequency residual correction in the debiased domain, *c.f.* Eqs. (11) and (12).

## D. Simulation of Poisson-Gaussian Noise

For our experiments on synthetic data (Sec. 5 of main paper) we apply Poisson-Gaussian noise to the noise-free images (Eq. 13). To demonstrate that Eq. (13) is sensible let $x'_n$ again be the unclipped noisy signal. According to the heteroscedastic Gaussian noise model, *c.f.* Eqs. (2a) and (2b), the mean and variance of $x'_n$ are given by

$$\mathbb{E}(x'_n) = y_n, \tag{38}$$

$$\text{Var}(x'_n) = \beta_1^n y_n + \beta_2^n. \tag{39}$$

Let now $z_n$ be the unclipped simulated noisy signal of Eq. (13):

$$z_n \sim \beta_1^n \cdot \mathcal{P}(y_n/\beta_1^n) + \mathcal{N}\big(0, \sqrt{\beta_2^n}\big) \tag{40}$$

According to the properties of the Poisson distribution, the mean and variance of $z_n$ are given by

$$\mathbb{E}(z_n) = \beta_1^n \frac{y_n}{\beta_1^n} + 0 = y_n, \tag{41}$$

$$\text{Var}(z_n) = (\beta_1^n)^2 \frac{y_n}{\beta_1^n} + \beta_2^n = \beta_1^n y_n + \beta_2^n. \tag{42}$$

We can see that the two first moments of $x'_n$ and $z_n$ match and hence $z_n$ provides a good simulation of the noise as given by the noise level function $\boldsymbol{\beta}^n$. The same holds for the simulation of the reference image.

## E. Additional Results

Finally, we give a few more results obtained on our novel DND benchmark dataset. First, Fig. 8 shows a histogram of the PSNR values of the crops of the noisy test images in linear raw space. As we can see, our dataset covers a wide range of noise levels for the noisy images, hence allowing to benchmark denoising algorithms across many different situations. Note that the mean PSNR of the noisy images (39.38 dB) is significantly below the PSNR of the reference images (52.76 dB, from the estimated noise level function). Consequently, the ground truth accuracy of our benchmark far exceeds the performance of state-of-the-art denoising techniques (*c.f.* Table 3), thus providing significant headroom even for future improvement in denoising techniques. Figure 9 shows denoising results aggregated for different noise levels. The top-performing methods overall achieve consistent results across almost all noise levels. We can furthermore observe that NCSR has severe problems in denoising images affected by weak intensity-dependent noise. When applying the variance stabilizing transformation, NCSR shows a more competitive performance. For MLP we observe that performance on RAW

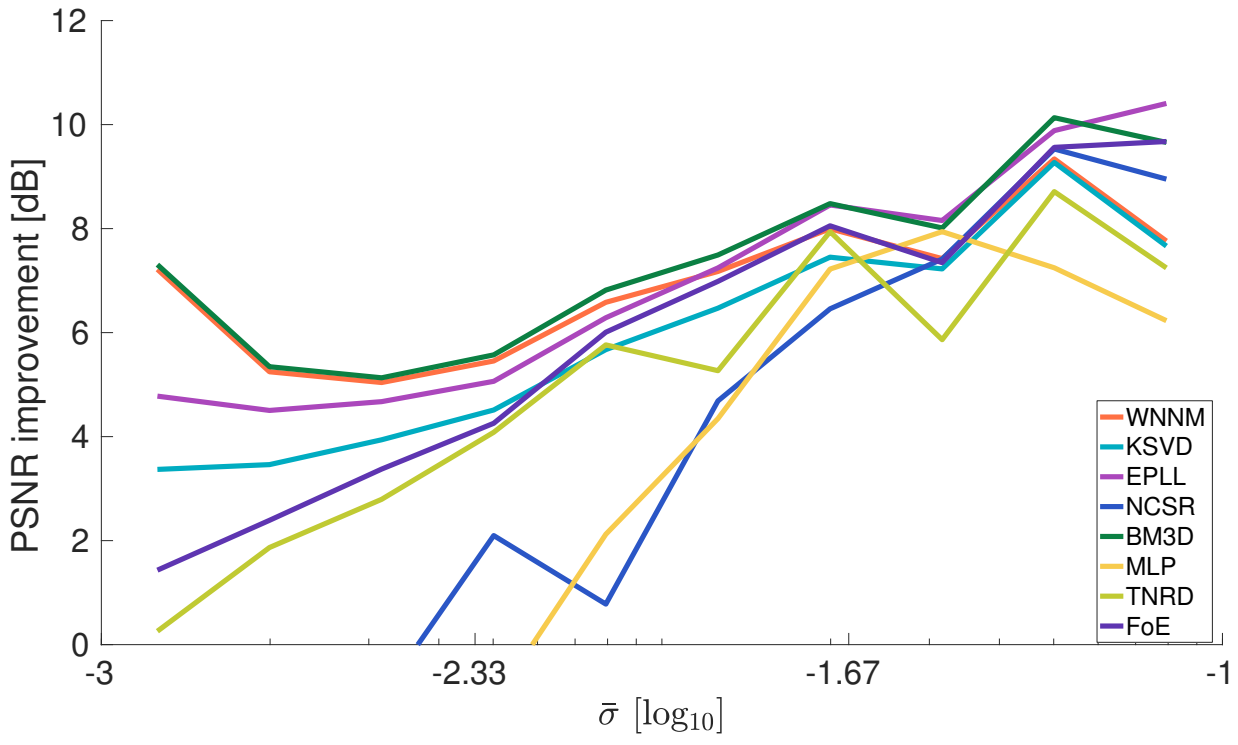

(a) Denoising raw pixels, evaluating in RAW space.

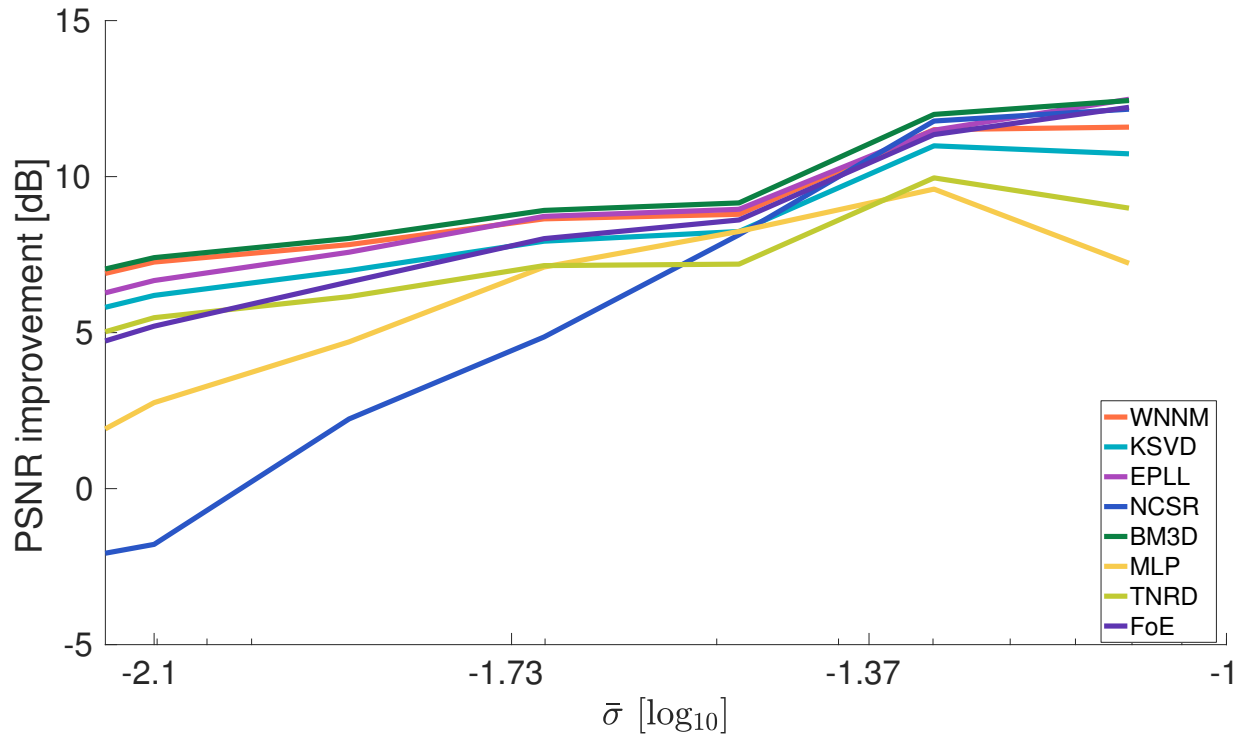

(b) Denoising raw pixels, evaluating in sRGB space.

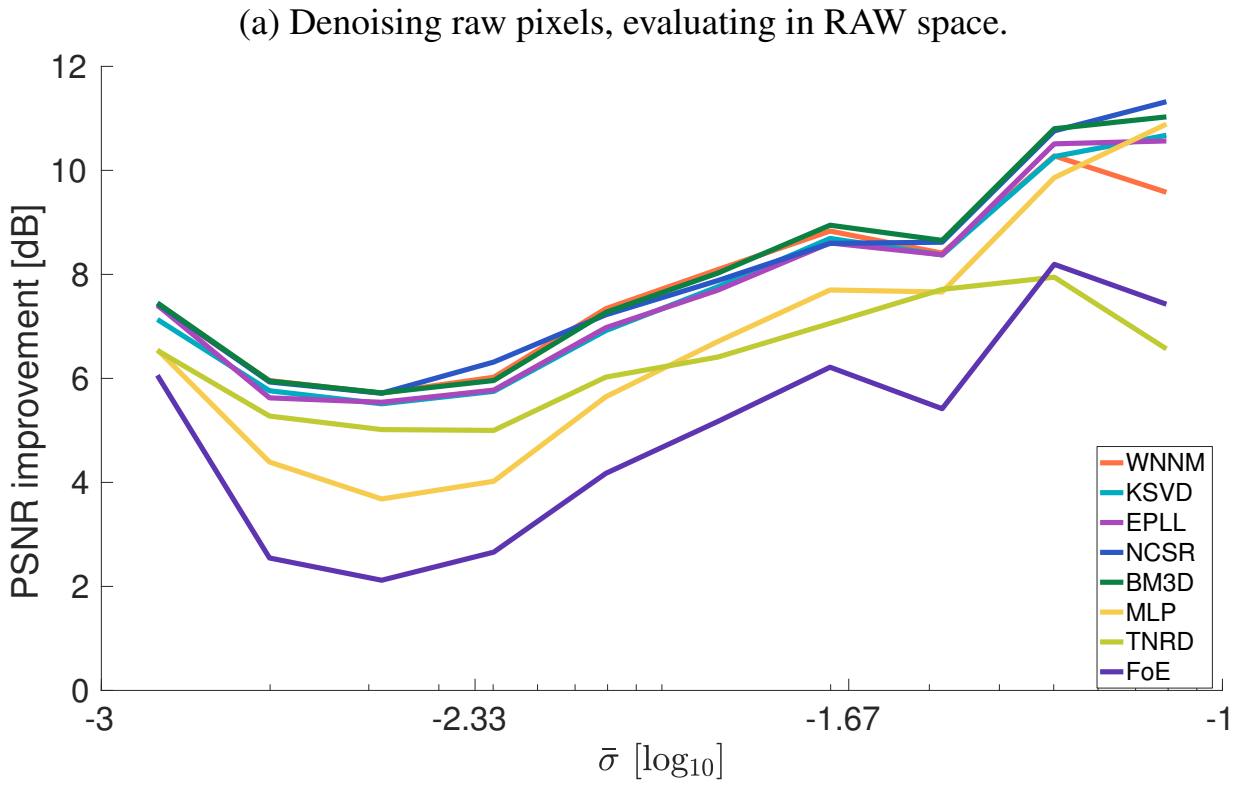

(c) Denoising raw pixels after VST, evaluating in RAW space.

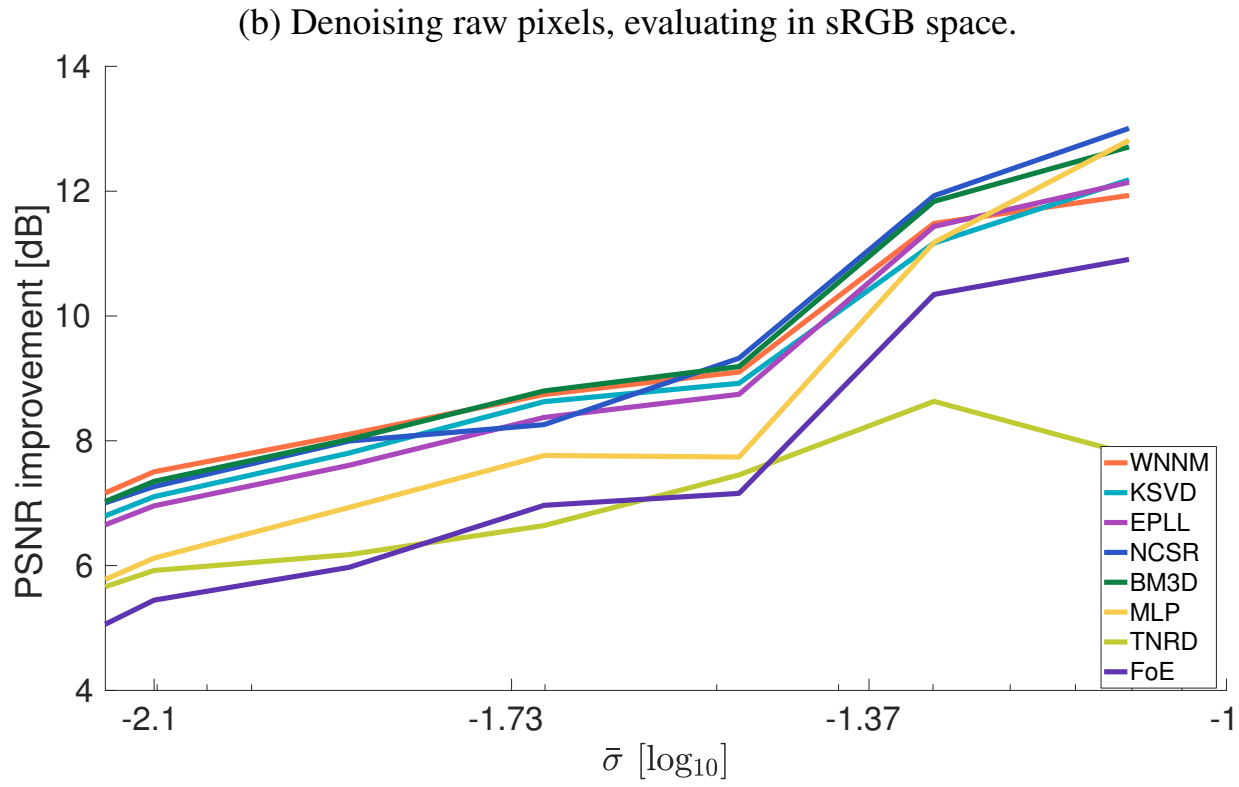

(d) Denoising raw pixels after VST, evaluating in sRGB space.

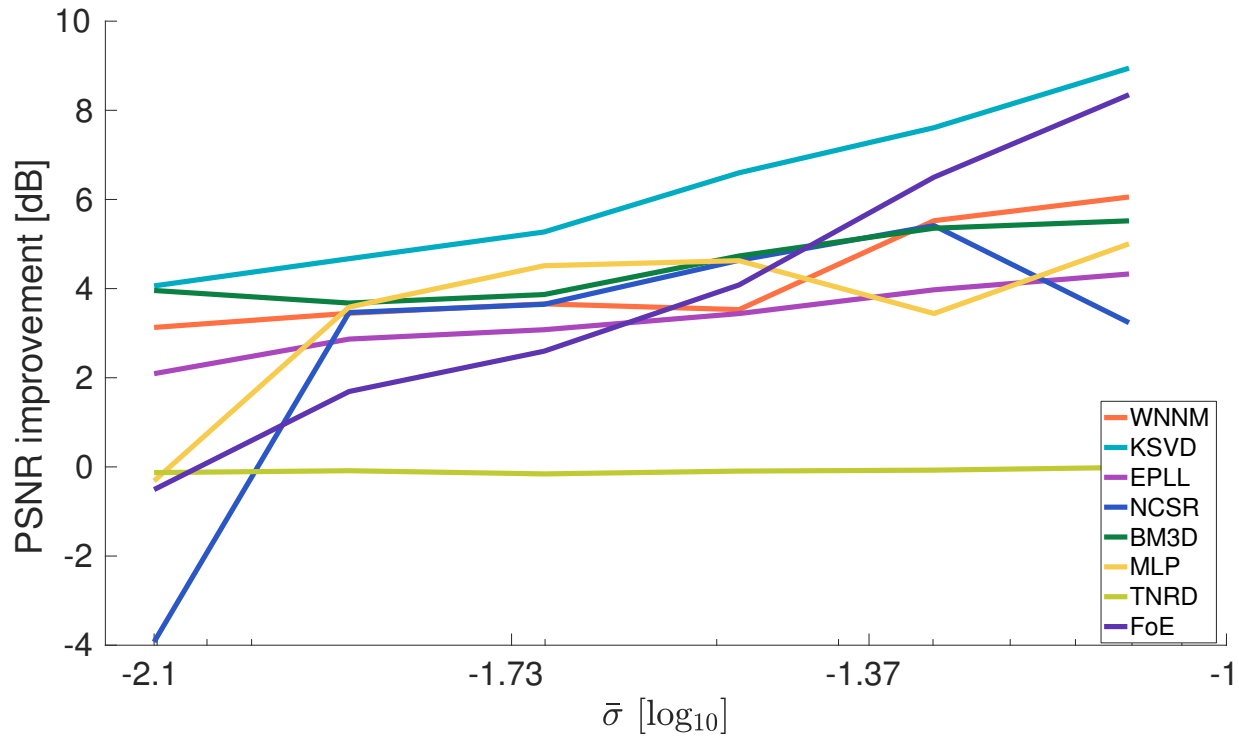

(e) Denoising in sRGB space, evaluating in sRGB space.

Figure 9. Denoising performance by noise level $\bar{\sigma}$.

denoising peaks for $\hat{\sigma}$ close to the noise level used for training, *i.e.* $\sigma_{\text{train}} \approx 10^{-1.41}$. For removing noise with a different noise level, MLP does not generalize well.

Table 4 provides SSIM [36] results for the tested methods on our benchmark. Generally, the conclusions made in Sec. 6 of the paper based on PSNR values generalize to the SSIM results. As we can see, BM3D and WNNM show the best performance and their scores differ only marginally. Overall, we observe that SSIM scores are high across all methods.

Finally, Figures 10 – 13 show denoising results of the tested algorithms for one crop of two different images in our database. The results were obtained from denoising raw intensities after the variance stabilizing transformation. We display the denoised images both in linear raw space (red channel only) and in sRGB space after our camera processing pipeline, *c.f.* Sec. 6 of the paper. On Figs. 10 and 12 we can see that many methods oversmooth fine structures (*e.g.*, MLP and FoE), while TNRD undersmoothes and fails to remove a significant part of the noise. When looking at the results in sRGB space (Figs. 11 and 13), we can see that denoising introduces visually apparent color artifacts for all methods. Moreover, the noise is clearly spatio-chromatically correlated in sRGB space.

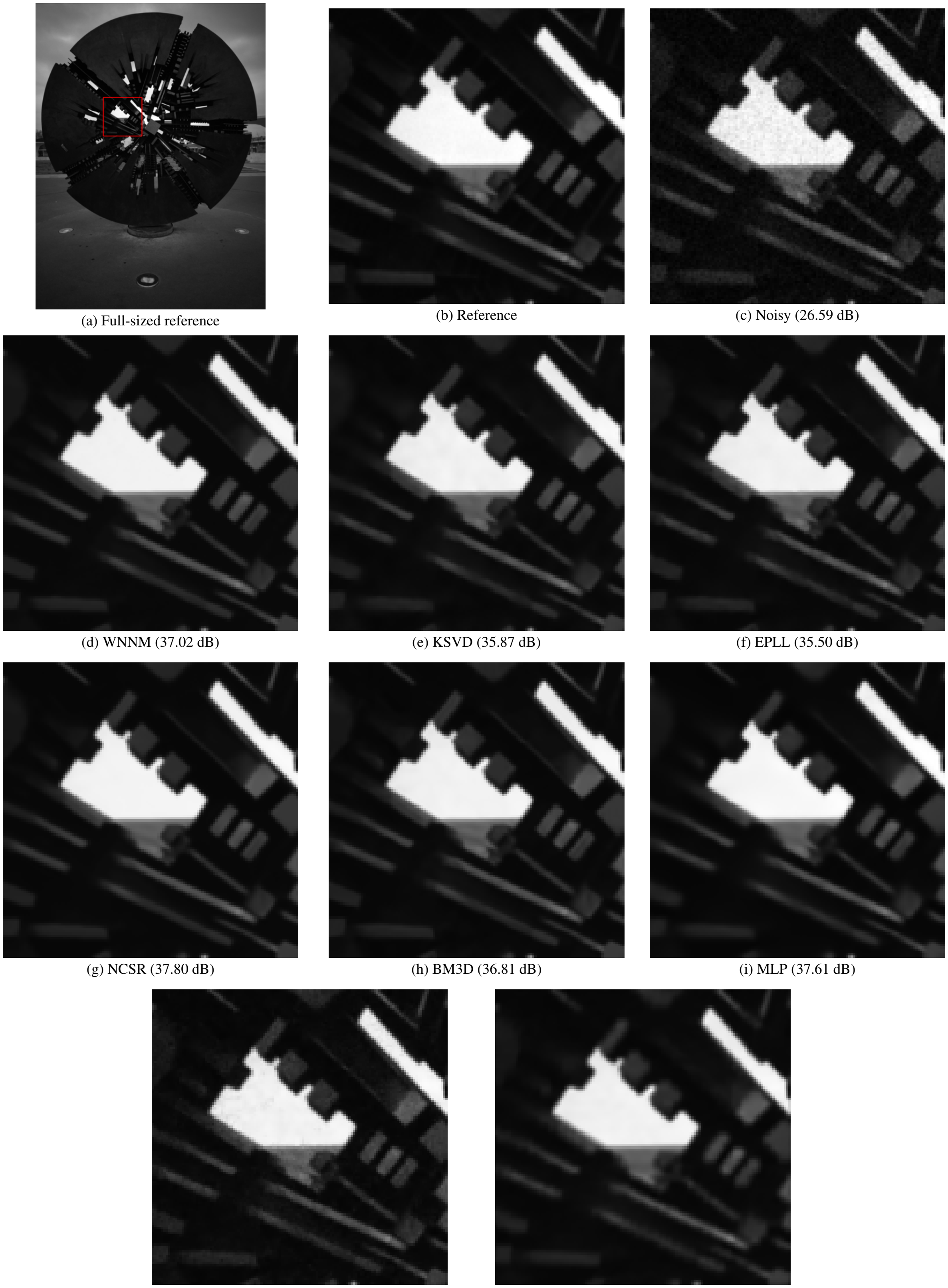

(a) Full-sized reference (b) Reference (c) Noisy (26.59 dB)
(d) WNNM (37.02 dB) (e) KSVD (35.87 dB) (f) EPLL (35.50 dB)
(g) NCSR (37.80 dB) (h) BM3D (36.81 dB) (i) MLP (37.61 dB)
(j) TNRD (32.38 dB) (k) FoE (32.39 dB)

Figure 10. Example denoising result (red channel only) with PSNR values, displayed in linear raw space.

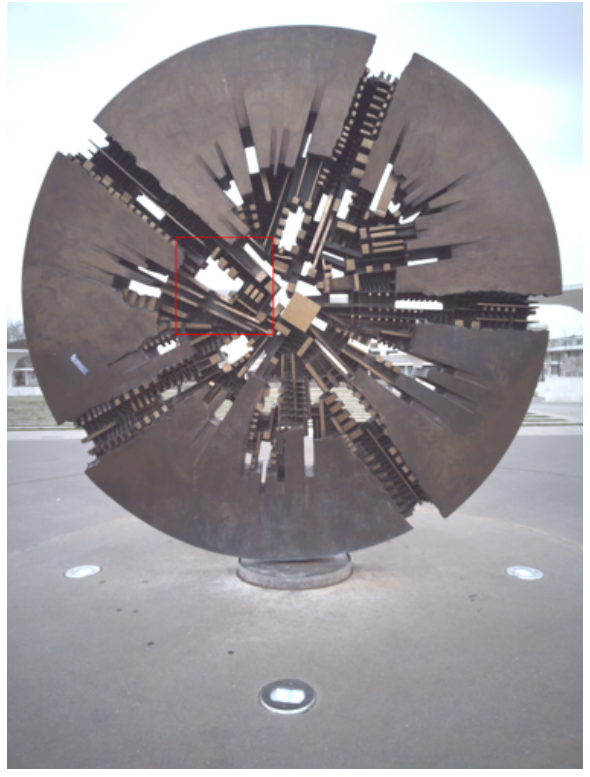

(a) Full-sized reference

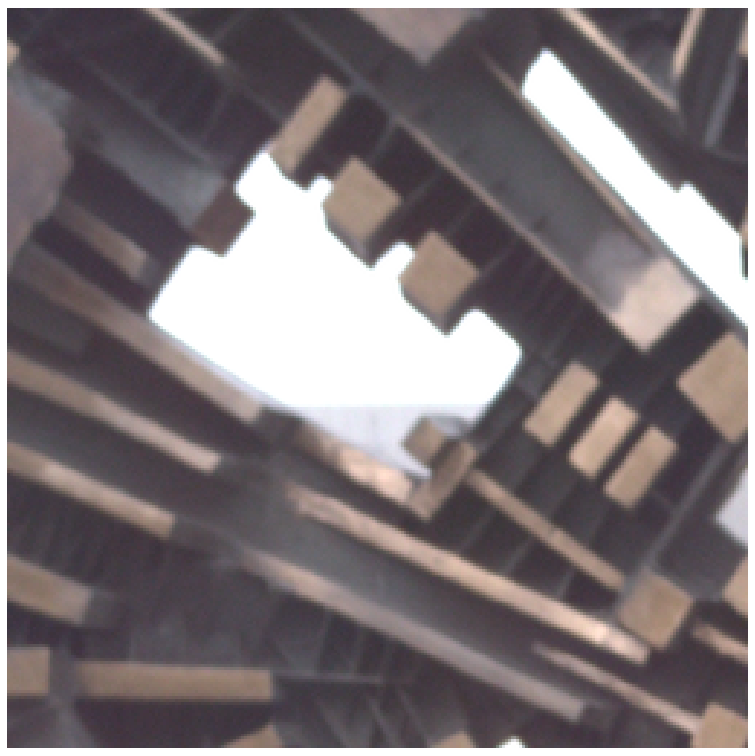

(b) Reference

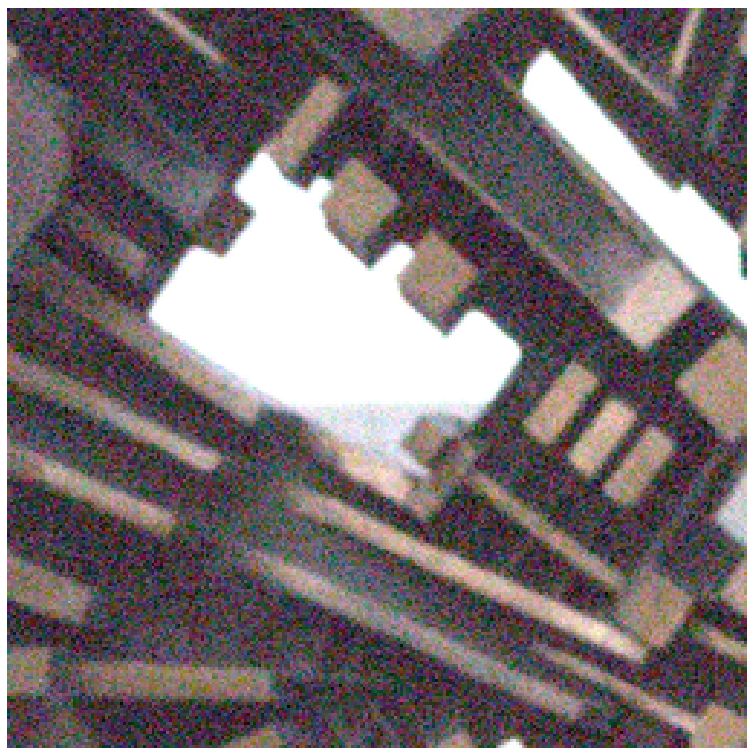

(c) Noisy (19.17 dB)

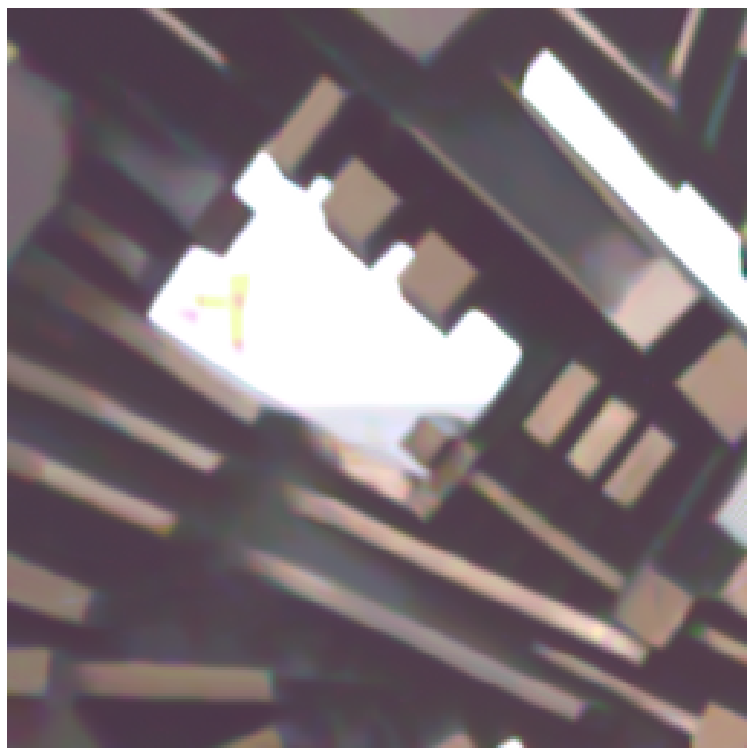

(d) WNNM (30.85 dB)

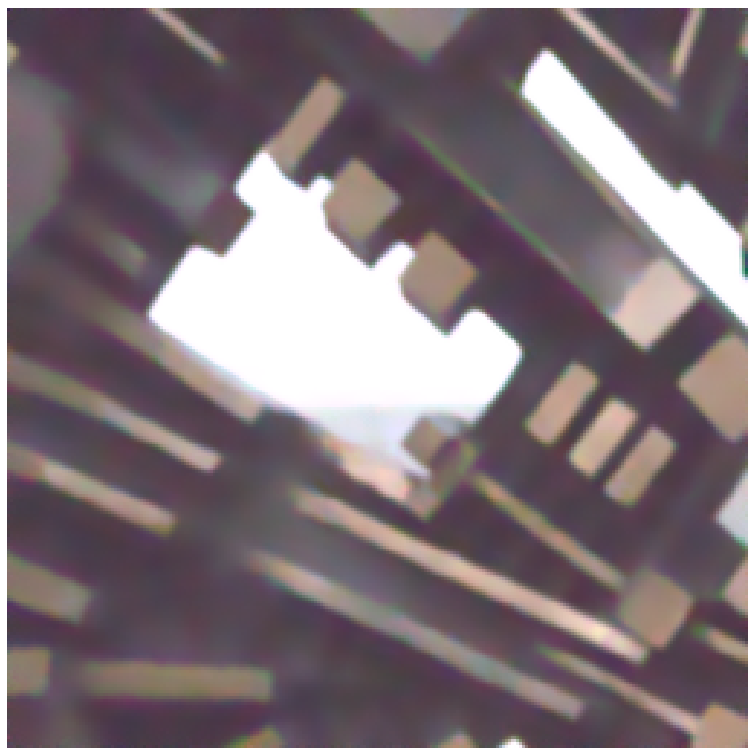

(e) KSVD (29.76 dB)

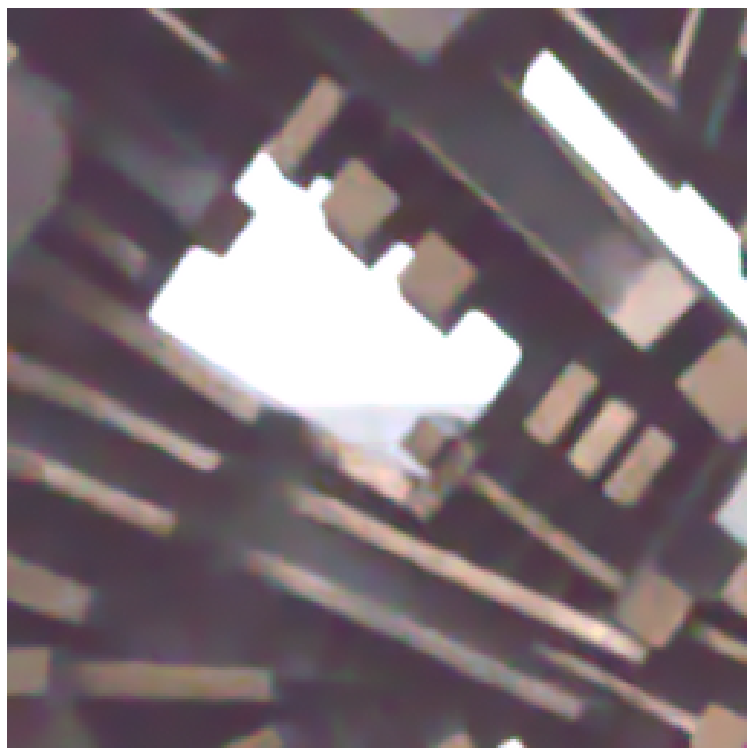

(f) EPLL (30.26 dB)

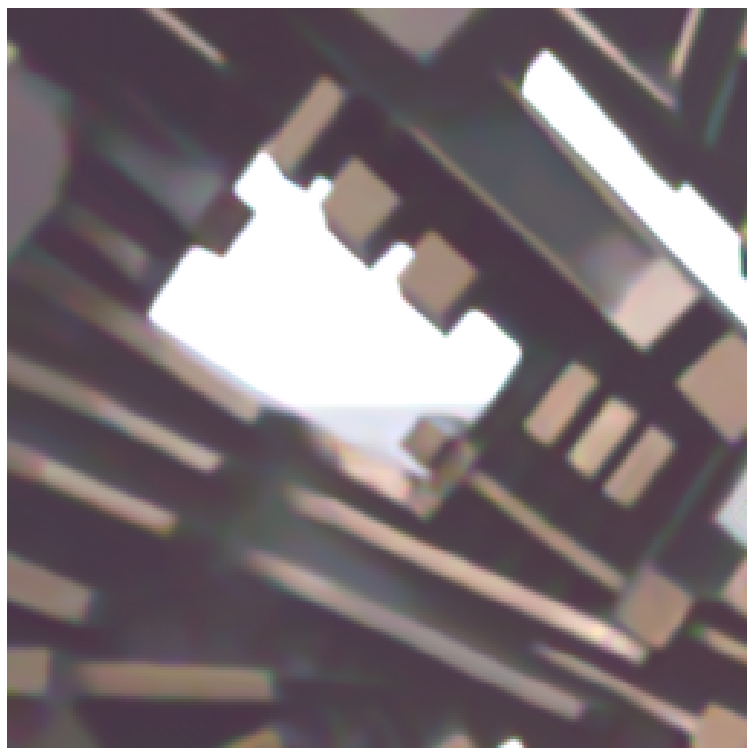

(g) NCSR (31.90 dB)

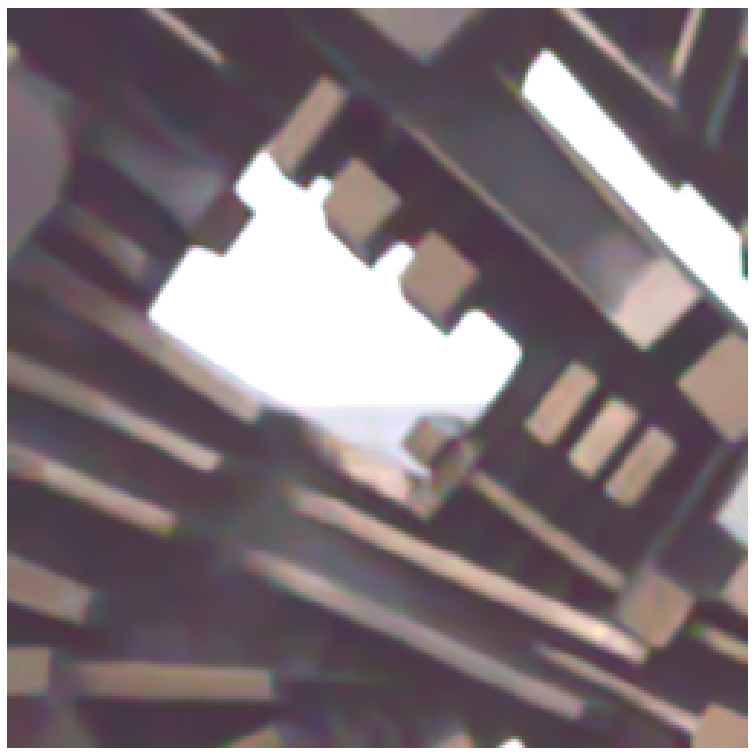

(h) BM3D (31.36 dB)

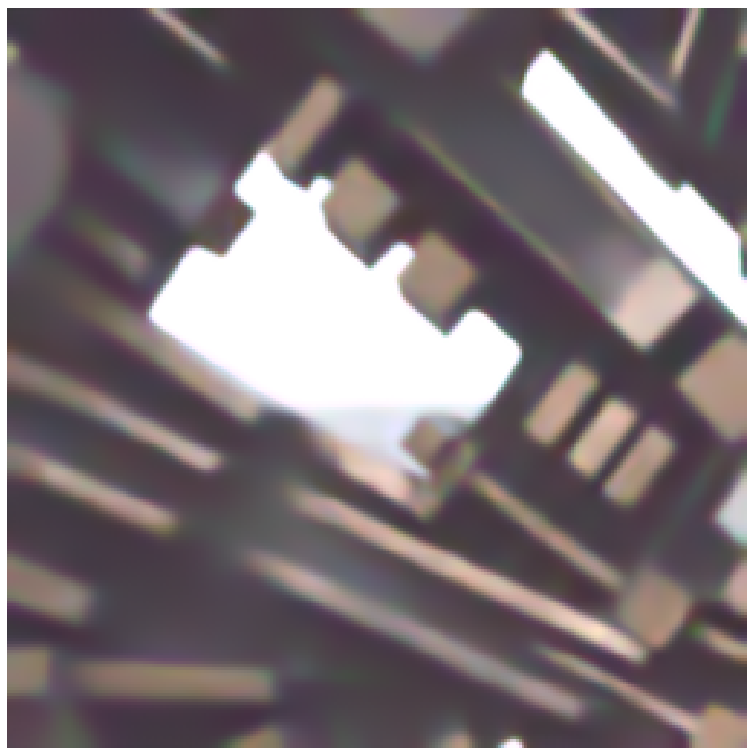

(i) MLP (31.38 dB)

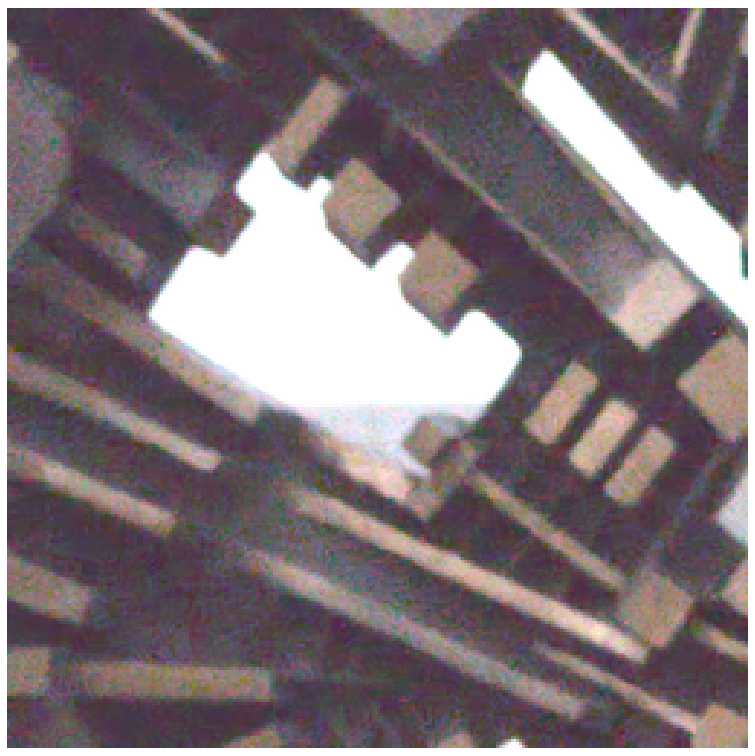

(j) TNRD (26.51 dB)

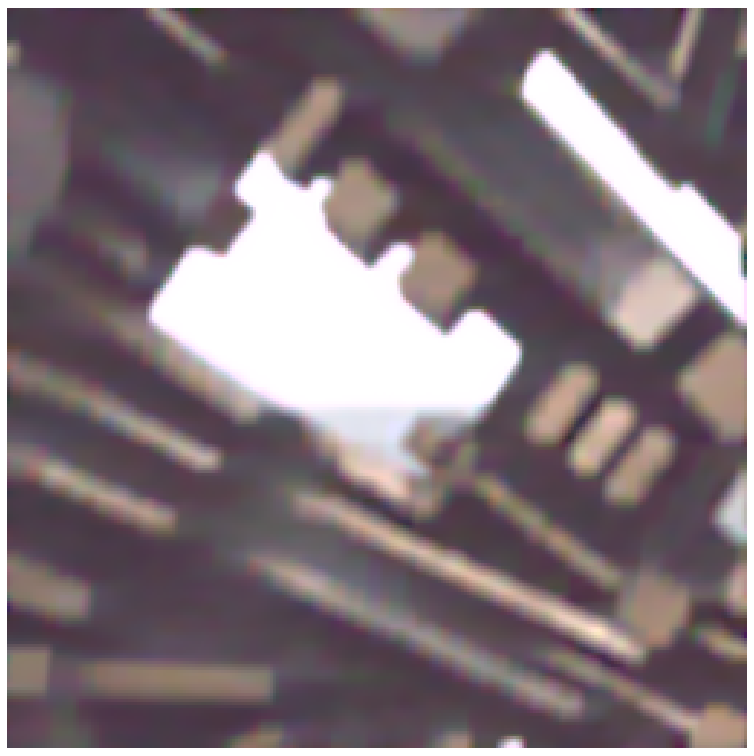

(k) FoE (27.81 dB)

Figure 11. Example denoising result (red channel only) with PSNR values, displayed in sRGB space.

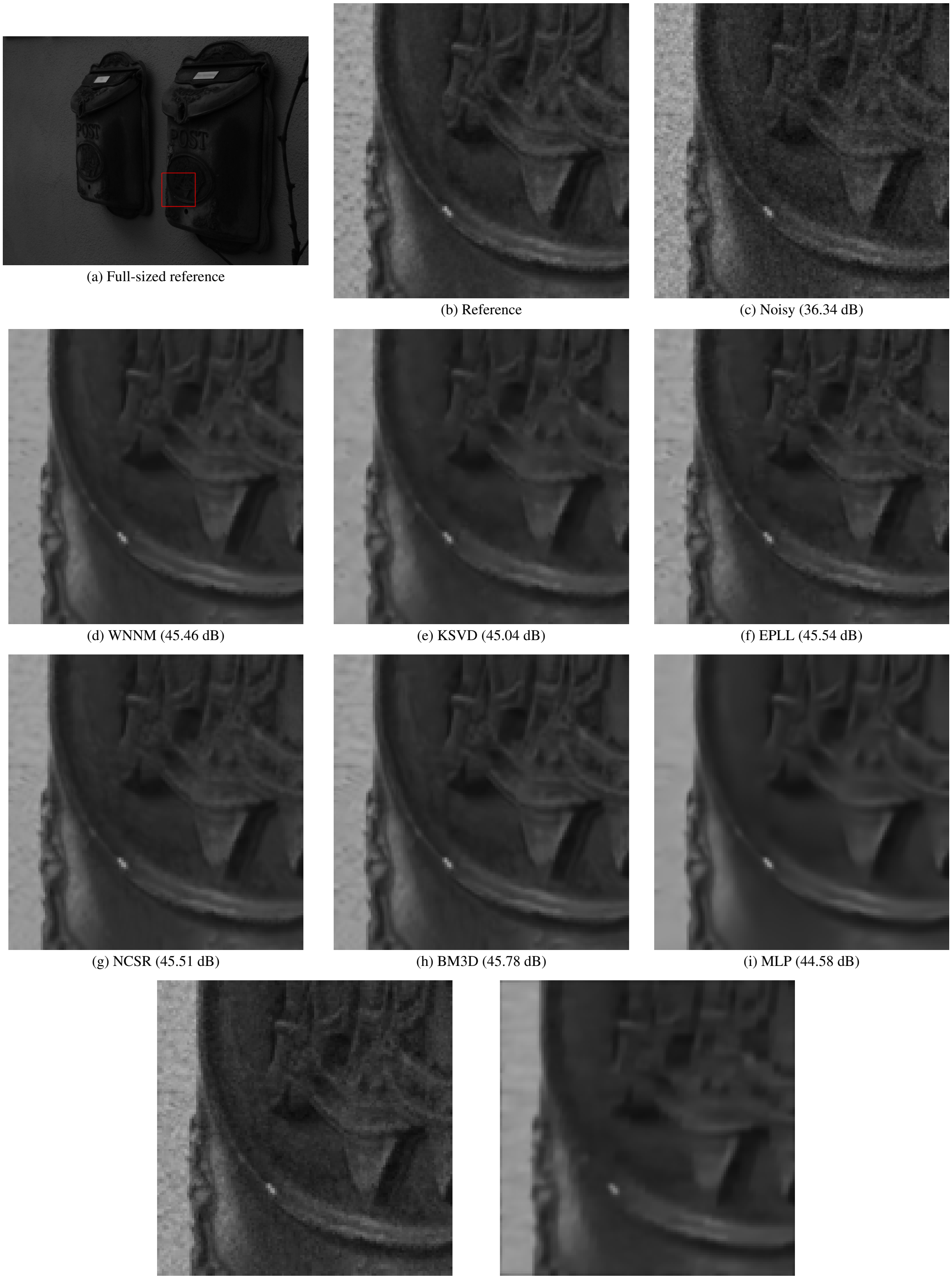

(a) Full-sized reference

(b) Reference

(c) Noisy (36.34 dB)

(d) WNNM (45.46 dB)

(e) KSVD (45.04 dB)

(f) EPLL (45.54 dB)

(g) NCSR (45.51 dB)

(h) BM3D (45.78 dB)

(i) MLP (44.58 dB)

(j) TNRD (40.44 dB)

(k) FoE (42.50 dB)

Figure 12. Example denoising result (red channel only) with PSNR values, displayed in linear raw space. Intensities of crops are uniformly scaled for better display.

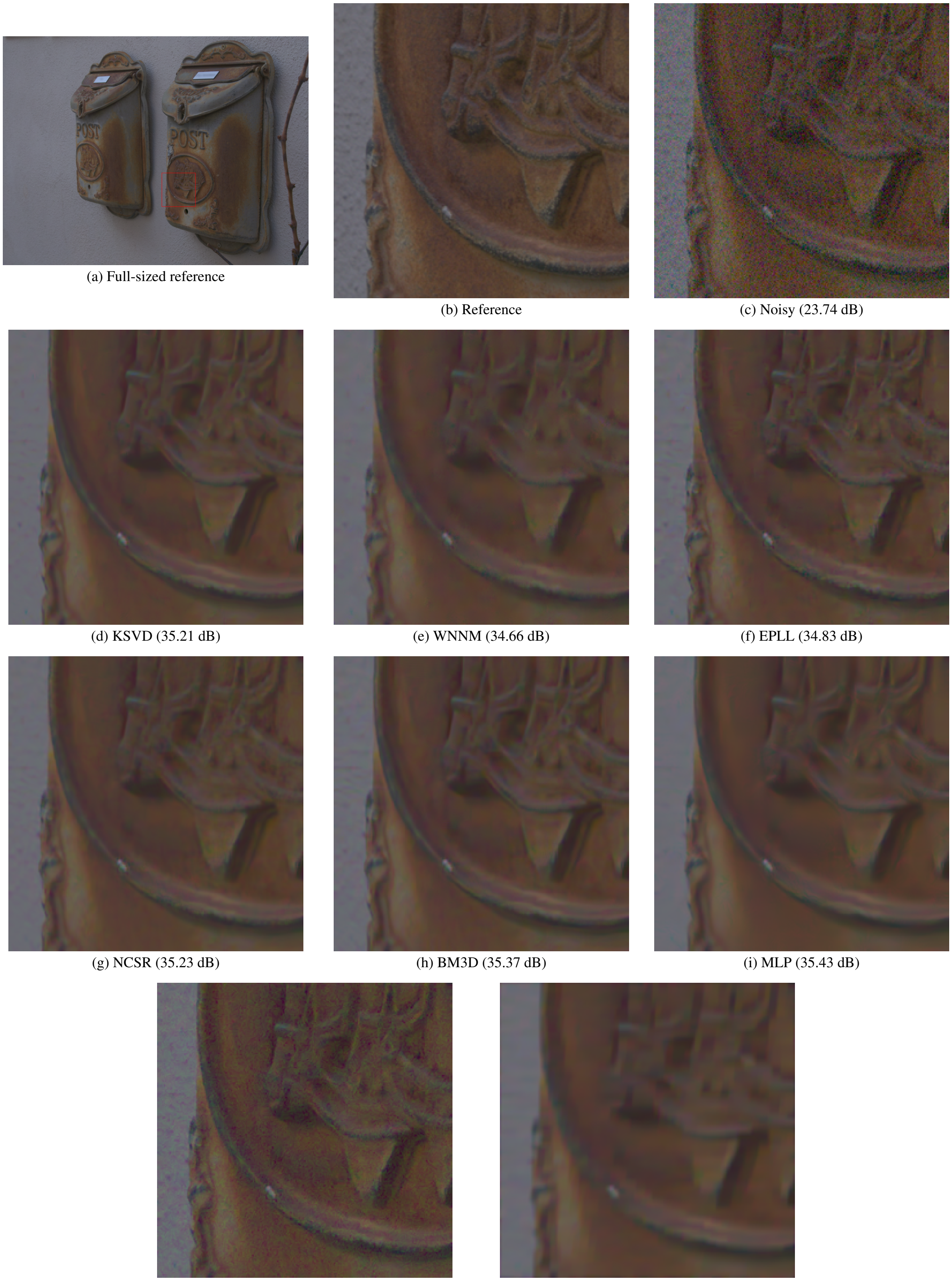

(a) Full-sized reference (b) Reference (c) Noisy (23.74 dB)

(d) KSVD (35.21 dB) (e) WNNM (34.66 dB) (f) EPLL (34.83 dB)

(g) NCSR (35.23 dB) (h) BM3D (35.37 dB) (i) MLP (35.43 dB)

(j) TNRD (31.94 dB) (k) FoE (34.00 dB)

Figure 13. Example denoising result with PSNR values, displayed in sRGB space.